\documentclass[12pt]{article}

\usepackage{paper-macros}
\usepackage{format-macros-arxiv}

\draftcompilefalse

\includestandalonefalse

\title{\LARGE \bfseries Optimal Mistake Bounds \\ for Transductive Online Learning}
\author{%
    \begin{minipage}[t]{0.45\textwidth}
        \centering
        {\normalsize Zachary Chase} \\
        {\small \textit{Kent State University}} \\
        {\small$\mathtt{zchase2@kent.edu}$}
    \end{minipage}
    \hfill
    \begin{minipage}[t]{0.45\textwidth}
        \centering
        {\normalsize Steve Hanneke} \\
        {\small \textit{Purdue University}} \\
        {\small$\mathtt{steve.hanneke@gmail.com}$}
    \end{minipage}
    \vspp \vsp \\
    \begin{minipage}[t]{0.45\textwidth}
        \centering
        {\normalsize Shay Moran} \\
        {\small \textit{Departments of Mathematics, Computer Science,}} \\
        {\small \textit{and Data and Decision Sciences}} \\
        {\small \textit{Technion -- Israel Institute of Technology;}} \\
        {\small \textit{Google Research}} \\
        {\small$\mathtt{smoran@technion.ac.il}$}
    \end{minipage}
    \hfill
    \begin{minipage}[t]{0.45\textwidth}
        \centering
        {\normalsize Jonathan Shafer} \\
        {\small \textit{MIT}} \\
        {\small$\mathtt{shaferjo@mit.edu}$}
    \end{minipage}
    \vspp
    \null\\[2.5em]
}

\date{\vspace*{-1em}}

\begin{document}

\maketitle

\pagenumbering{gobble}

    \ifdraftcompile
        \begin{abstract}
    We resolve a 30-year-old open problem concerning the power of unlabeled data in online learning by tightly quantifying the gap between transductive and standard online learning. 
    In the standard setting, the optimal mistake bound is characterized by the Littlestone dimension $d$ of the concept class $\mathcal{H}$~\citep*{DBLP:journals/ml/Littlestone87}. 
    We prove that in the transductive setting, the mistake bound is at least $\Omegasqrtd$. 
    This constitutes an exponential improvement over previous lower bounds of $\OmegaOf{\log\log d}$, $\Omegasqrtlogd$, and $\OmegaOf{\log d}$, due respectively to \cite*{DBLP:conf/eurocolt/Ben-DavidKM95,DBLP:journals/ml/Ben-DavidKM97}, and \cite*{DBLP:conf/nips/HannekeMS23}.
    We also show that this lower bound is tight: for every~$d$, there exists a class of Littlestone dimension~$d$ with transductive mistake bound $\Osqrtd$. 
    Our upper bound also improves upon the best known upper bound of $(2/3)\cdot d$ from \cite{DBLP:journals/ml/Ben-DavidKM97}. 
    These results establish a quadratic gap between transductive and standard online learning, thereby highlighting the benefit of advance access to the unlabeled instance sequence. 
    This contrasts with the PAC setting, where transductive and standard learning exhibit similar sample complexities.
\end{abstract}

    \else
        \begin{abstract}
    We resolve a 30-year-old open problem concerning the power of unlabeled data in online learning by tightly quantifying the gap between transductive and standard online learning. 
    In the standard setting, the optimal mistake bound is characterized by the Littlestone dimension $d$ of the concept class $\mathcal{H}$~\citep*{DBLP:journals/ml/Littlestone87}. 
    We prove that in the transductive setting, the mistake bound is at least $\Omegasqrtd$. 
    This constitutes an exponential improvement over previous lower bounds of $\OmegaOf{\log\log d}$, $\Omegasqrtlogd$, and $\OmegaOf{\log d}$, due respectively to \cite*{DBLP:conf/eurocolt/Ben-DavidKM95,DBLP:journals/ml/Ben-DavidKM97}, and \cite*{DBLP:conf/nips/HannekeMS23}.
    We also show that this lower bound is tight: for every~$d$, there exists a class of Littlestone dimension~$d$ with transductive mistake bound $\Osqrtd$. 
    Our upper bound also improves upon the best known upper bound of $(2/3)\cdot d$ from \cite{DBLP:journals/ml/Ben-DavidKM97}. 
    These results establish a quadratic gap between transductive and standard online learning, thereby highlighting the benefit of advance access to the unlabeled instance sequence. 
    This contrasts with the PAC setting, where transductive and standard learning exhibit similar sample complexities.
\end{abstract}

    \fi

\newpage

\tableofcontents

\newpage

\pagenumbering{arabic}

    \ifdraftcompile
        \section{Introduction}

The transductive model is a basic and well-studied framework in learning theory, dating back to the early works of Vapnik. It has been investigated both in statistical and online settings, and is motivated by the principle that to make good predictions on a specific set of test instances, one need not construct a fully general classifier that performs well on the entire domain --- including points that may never actually appear. Rather, it may be sufficient to tailor predictions for a fixed, known set of instances.

This perspective naturally connects to a broader question in learning theory: what is the value of unlabeled data? In the transductive setting, the learner is given the sequence of unlabeled test instances in advance and is then required to predict their labels one by one. Thus, the transductive model can be viewed as a natural formalization of learning with unlabeled data: the test instances are known in advance, but their labels are not. The central question is whether such prior access to the unlabeled sequence can help reduce the number of prediction mistakes --- compared to the standard online model, where the instances arrive and are labeled one at a time.

Recall for instance that in the standard PAC\footnote{%
    Probably Approximately Correct. For an exposition of the standard terminology and results mentioned in this paragraph see, e.g., \cite{DBLP:books/daglib/0033642}.
} model of supervised learning, there are cases where access to unlabeled data is not helpful. Indeed, the ``hard population distributions'' used to prove the standard VC\footnote{%
    Vapnik--Chervonenkis.
} lower bound are constructed by taking a fixed and known marginal distribution over a VC-shattered set. Namely, the cases that are hardest to learn in the PAC setting include ones where the learner knows the marginal distribution over the domain, and can therefore generate as much unlabeled data as it wishes. And yet, in those cases, access to unlabeled data provides no acceleration compared to an algorithm (like ERM\footnote{Empirical Risk Minimization.}) that does not use unlabeled data.

Seeing as unlabeled data is often a lot easier to obtain than labeled data, there have been considerable efforts to understand when and to what extent can access to unlabeled data accelerate learning.\footnote{%
    The literature on semi-supervised learning is surveyed in \cite*{DBLP:conf/icml/Joachims99,zhu2005semi,%
    DBLP:series/synthesis/2009Zhu,%
    DBLP:reference/ml/Zhu10,%
    DBLP:books/mit/06/CSZ2006}. Theoretical works on the topic include \cite*{DBLP:journals/tcs/BenedekI91,%
    DBLP:conf/colt/BlumM98,%
    DBLP:journals/corr/abs-0805-2891,%
    DBLP:journals/jacm/BalcanB10,%
    DBLP:conf/stacs/DarnstadtSS13,%
    DBLP:conf/colt/GopfertBBGTU19}.
}

In particular, it is natural to ask, for which plausible models of learning is access to unlabeled data beneficial? Online learning \citep{DBLP:journals/ml/Littlestone87} is perhaps the model of learning that is most-extensively studied in learning theory after the PAC model and its variants. Therefore, the general question considered in this paper is:

\vsp

\begin{question}
    \label{question:general}
    Quantitatively, how much (if at all) is access to unlabeled data beneficial for learning in the online learning setting?
\end{question}

This question is naturally instantiated by comparing \emph{transductive} online learning --- where the learner has advance access to the full sequence $x_1,x_2,\dots,x_n$ of unlabeled instances --- with \emph{standard} online learning, where no such access is given. This perspective has also been adopted in prior work: for example, \cite{DBLP:conf/nips/KakadeK05}, \cite{DBLP:conf/birthday/Cesa-BianchiS13}, and \cite*{DBLP:journals/ijon/HoiSLZ21} (Section 7.3) all describe transductive online learning as a setting in which the learner has access to ``unlabeled data''. We thus refine the question above as follows:

\vsp

\begin{question}
    \label{question:concrete}
    Quantitatively, how much (if at all) is learning in the transductive online learning setting easier than learning in the standard online learning setting? Specifically, how much is the optimal number of mistakes in the transductive setting smaller than in the standard setting?
\end{question}

Addressing this question, our main result (\cref{theorem:main-result}) states that the optimal number of mistakes in the transductive setting (with access to unlabeled data) is at most quadratically smaller than in the standard setting (without unlabeled data). Furthermore, there are hypothesis classes for which a quadratic gap is achieved.

\subsection{Setting: \emph{Standard} vs.\ \emph{Transductive} Online Learning}
\label{section:setting}

\emph{Standard online learning} \citep{DBLP:journals/ml/Littlestone87} is a zero-sum, perfect- and complete-information game played over $n$ rounds between two players, a \emph{learner} and an \emph{adversary}. The game is played with respect to a \emph{domain} set $\cX$ and a \emph{hypothesis class} $\cH \subseteq \zo^{\cX}$ (consisting of functions $\cX \to \zo$), where $n$, $\cX$ and $\cH$ are fixed and known to both players. The game proceeds as \ifneurips{in \cref{game:standard}}\else{follows}\fi.
\ifneurips\begin{game}\else\begin{game}[H]\fi
    \begin{ShadedBox}
        For each round $t = 1,2,\dots,n$:
        \vsp
        \begin{enumerate}[label=\textit{\alph*}.]
            \item{
                The adversary selects an \emph{instance} $x_t \in \cX$ and sends it to the learner.
            }
            \vsp
            \item{
                The learner selects a \emph{prediction} $\hat{y}_t \in \zo$ and sends it to the adversary.
            }
            \vsp
            \item{
                The adversary selects a \emph{label} $y_t \in \zo$ and sends it to the learner. The selected label must be \emph{realizable}, meaning that $\exists h \in \cH ~ \forall i \in [t]\!: ~ h(x_i) = y_i$. 
            }
        \end{enumerate}
    \end{ShadedBox}
    \vspace{-1em}
    \caption{The standard online learning setting.}
    \label{game:standard}
\end{game}
\ifneurips{}\else{\vspace*{-1em}}\fi
The \emph{number of mistakes} for a learner $L$ and an adversary $A$ is $\onlinemistakes{\cH,n,\learner,\adversary} = \abs*{\set*{t \in [n]: ~ \hat{y}_t \neq y_t}}$. We are interested in understanding the \emph{optimal number of mistakes}, which is 
\[
    \onlinemistakes{\cH} = 
    \sup_{n \in \bbN} 
    \: 
    \inf_{\learner \in \learners}
    \:
    \sup_{\adversary \in \adversaries} 
    \: 
    \onlinemistakes{\cH,n,\learner,\adversary},
\]
where $\adversaries$ and $\learners$ are the set of all deterministic adversaries and learners, respectively.\footnote{%
    In this paper, we only consider deterministic learners and adversaries. As is common in learning theory, we avoid questions of computability and allow the learner and adversary to be any function. See \cref{section:preliminaries} for formal definitions of $\adversaries$ and $\learners$.
    
    One could consider an alternative setting where the learner and adversary are randomized and the adversary selects each label $y_t$ \emph{before} the learner tosses the random coins to determine the prediction $\hat{y}_t$; the optimal number of mistakes in the alternative setting equals the optimal number of mistakes in our deterministic setting up to a factor of two.%
}\ifneurips{\par}\fi{ }It is well known that $\onlinemistakes{\cH}$ is characterized by the Littlestone dimension, namely, $\onlinemistakes{\cH} = \LD{\cH}$ (see \cref{theorem:ld-is-upper-bound,definition:littlestone-tree-and-dimension}).

The \emph{transductive} online learning setting \citep{DBLP:conf/eurocolt/Ben-DavidKM95,DBLP:journals/ml/Ben-DavidKM97} is similar, except that the learner has access to the full sequence of unlabeled instances in advance. Namely, \ifneurips{as in \cref{game:transductive}.}\fi
\ifneurips\begin{game}\else\begin{game}[H]\fi
    \begin{ShadedBox}
        The adversary selects a \emph{sequence} $x_1,x_2,\dots,x_n \in \cX$ and sends it to the learner.

        \vsp
        
        For each round $t = 1,2,\dots,n$:
        \vsp
        \begin{enumerate}[label=\textit{\alph*}.]
            \item{
                The learner selects a \emph{prediction} $\hat{y}_t \in \zo$ and sends it to the adversary.
            }
            \vsp
            \item{
                The adversary selects a \emph{label} $y_t \in \zo$ and sends it to the learner. The selected label must be \emph{realizable}, meaning that $\exists h \in \cH ~ \forall i \in [t]\!: ~ h(x_i) = y_i$. 
            }
        \end{enumerate}
    \end{ShadedBox}
    \vspace{-1em}
    \caption{The transductive online learning setting.}
    \label{game:transductive}
\end{game}
\ifneurips{}\else{\vspace*{-1em}}\fi
The optimal number of mistakes for the transductive setting is defined exactly as before,
\begin{equation*}
    \transductivemistakes{\cH,n,\learner,\adversary} = \abs*{\set*{t \in [n]: ~ \hat{y}_t \neq y_t}},
    ~~
    \text{and}
    ~~
    \transductivemistakes{\cH} = 
    \sup_{n \in \bbN} 
    \: 
    \inf_{\learner \in \learners}
    \:
    \sup_{\adversary \in \adversaries} 
    \: 
    \transductivemistakes{\cH,n,\learner,\adversary},
\end{equation*}
with the only difference between the standard quantity $\onlinemistakes{\cH}$ and the transductive quantity $\transductivemistakes{\cH}$ being in how the game is defined.

\subsection{Main Result}

Notice that for every hypothesis class $\cH$, $\transductivemistakes{\cH} \leq \onlinemistakes{\cH}$. Indeed, in the transductive setting the adversary declares the sequence $x$ at the start of the game. This reduces the number of mistakes because the transductive adversary is less powerful (it cannot adaptively alter the sequence mid-game), and also because the transductive learner is more powerful (it has more information).\footnote{%
    One could also define an intermediate setting, where the adversary is less powerful because it must select the sequence at the start of the game and cannot change it during the gameplay, but the learner does not have more information because the adversary only reveals the instances in the sequence one at a time as in the standard setting. However, this intermediate setting would not model the learner having \emph{access} to unlabeled data.
    
    On a technical level, the intermediate setting is essentially not easier for the learner than the standard online learning setting. If the hypothesis class shatters a Littlestone tree of depth $d \in \bbN$, then for every deterministic learner there exists a deterministic adversary that forces the learner to make at least $d$ mistakes in the intermediate-transductive setting: The intermediate-transductive adversary simulates an execution of the deterministic learner playing against the standard online deterministic adversary in the \emph{standard} online learning setting to construct a sequence of instances $x_i$ and labels $y_i$ (corresponding to a root-to-leaf branch of the tree). After the simulation is complete, the intermediate-transductive adversary commits to that sequence of instances at the beginning of the intermediate-transductive game, and assigns the same labels that the adversary assigned in the simulation. Because the learner is deterministic and its view in the two games is the same, it is guaranteed to make at least $d$ mistakes on this sequence also in the intermediate-transductive setting.

    Similarly, if the learner is randomized and at each round the adversary selects a label before the learner tosses its random coins, then the intermediate-transductive adversary can force an expected number of $d/2$ mistakes by selecting a root-to-leaf branch of the tree uniformly at random.
}

While for some classes $\transductivemistakes{\cH} = \onlinemistakes{\cH}$, we study the largest possible separation. 
The best previous lower bound on $\transductivemistakesname$, due to \cite*{DBLP:conf/nips/HannekeMS23}, states that for every class~$\cH$,
\[
    \transductivemistakes{\cH} \geq \OmegaOf{\log d},
\]
where $d = \onlinemistakes{\cH}$.
In the other direction, \citet{DBLP:journals/ml/Ben-DavidKM97} constructed\footnote{%
    Their class consists of all disjoint unions of $\ThetaOf{d}$ functions from a specific constant-sized class.%
} a class $\cH$ such that $\onlinemistakes{\cH} = d$ and \ifneurips{$\transductivemistakes{\cH} \leq \tfrac{2}{3}d$.\nobreakspace}\else
\[
    \transductivemistakes{\cH}
    \leq \tfrac{2}{3}d.
\]
\fi{}This left an exponential gap between the best known lower and upper bounds on $\transductivemistakesname$, namely $\Omega(\log d)$ versus $\frac{2}{3}d$. Our main result closes this gap:

\begin{theorem}[Main result]
    \label{theorem:main-result}~
    \begin{itemize}
    \item{
        For every hypothesis class $\cH \subseteq \zo^{\cX}$,
        \[
        \transductivemistakes{\cH} = \Omegasqrtd,
        \]
        where $d = \onlinemistakes{\cH}.$
    }
    \item{
        On the other hand, for every $d$ there exists a hypothesis class $\cH$ with $\onlinemistakes{\cH} = d$ and 
        \[
            \transductivemistakes{\cH} = \Osqrtd.
        \]
    }
    \end{itemize}
\end{theorem}

This result is stated in considerably greater detail in \cref{theorem:lower-bound,theorem:separation}.

\subsection{Related Works}

The notion of \emph{transductive inference} as a more efficient alternative to \emph{inductive inference} in statistical learning theory was introduced by \cite*{vapnik1979estimation,DBLP:books/sp/Vapnik06,DBLP:conf/uai/GammermanAV98,DBLP:conf/nips/ChapelleVW99}. The \emph{online learning} setting is due to \cite*{DBLP:journals/ml/Littlestone87}, who also proved that the optimal number of mistakes is characterized by the Littlestone dimension (see \cref{theorem:ld-is-upper-bound}).

The \emph{transductive online learning} setting studied in the current paper, was first defined by \cite*{DBLP:conf/eurocolt/Ben-DavidKM95}, who used the name \emph{worst sequence off-line model}. Among other results, they showed a lower bound of $\OmegaOf{\log\log d}$ on the number of mistakes required to learn a class with Littlestone dimension $d$. The authors subsequently presented an exponentially stronger lower bound of $\Omegasqrtlogd$ in \cite*{DBLP:journals/ml/Ben-DavidKM97}. However, understanding where the optimal number of mistakes is situated within the range $\bracket*{\Omegasqrtlogd,2d/3}$ remained an open question.

\cite*{DBLP:conf/nips/KakadeK05} presented an oracle-efficient algorithm for the transductive online learning setting, and may have been the first to use that name. Their result was subsequently improved upon by \cite*{DBLP:conf/birthday/Cesa-BianchiS13}. 

The present work is most similar to that of \cite*{DBLP:conf/nips/HannekeMS23} which, among other results, gave a quadratically-stronger mistake lower bound of $\OmegaOf{\log d}$ for classes with Littlestone dimension $d$ in the transductive online setting. The proof of our lower bound utilizes some of their ideas, but yields a quantitative improvement by combining it with some new ideas.

\cite*{DBLP:conf/nips/HannekeRSS24} studied a setting of \emph{multi-class} transductive online learning where the number of possible labels is unbounded.

\begin{figure}[!ht]
    \centering
    \begin{subfigure}[T]{0.48\textwidth}
        \centering
        \includegraphicsorstandalone[width=\smalltreewidth\linewidth]{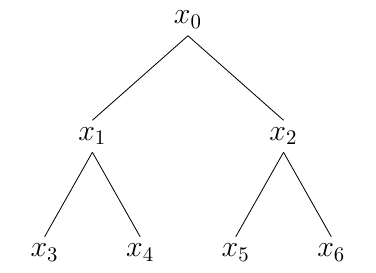}
        \vspace*{3.4em}
        \caption{
            A \emph{perfect binary tree} of depth $2$. Each \emph{node} is labeled by an element of the domain $\cX$. These labels need not be distinct (e.g., it is possible that $x_1 = x_6$). $x_0$ is the \emph{root} of the tree, $x_0$, $x_1$ and $x_2$ are \emph{internal nodes}, and $x_3,\dots,x_6$ are \emph{leaves}.
        }
        \label{subfigure:tree-basic}
    \end{subfigure}%
    \hfill
    \begin{subfigure}[T]{0.48\textwidth}
        \centering
        \includegraphicsorstandalone[width=\largetreewidth\linewidth]{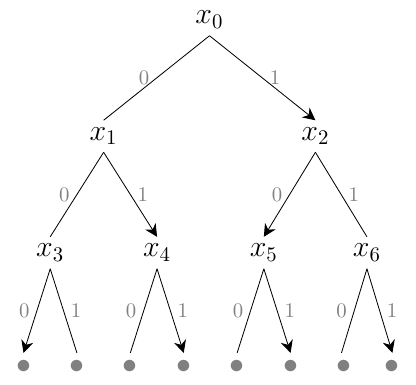}
        \caption{A function $f : ~ \cX \to \zo$ assigns a binary label to each node in the tree, represented here by edges with arrowhead tips. This figure depicts the function $f(x_i) = \indicator{i \notin \set{2,3}}$. (Note that the gray dots ({\color{gray}\ifneurips{\color{gray}$\bullet$}\else\raisebox{-0.2em}{\scalebox{2}{$\bullet$}}\fi}) in the figure are purely a pictorial detail. In this paper they are not considered nodes or leaves of the tree.)}
        \label{subfigure:tree-function}
    \end{subfigure}
    
    \vspace{1em}

    \begin{subfigure}[T]{0.48\textwidth}
        \centering
        \includegraphicsorstandalone[width=\largetreewidth\linewidth]{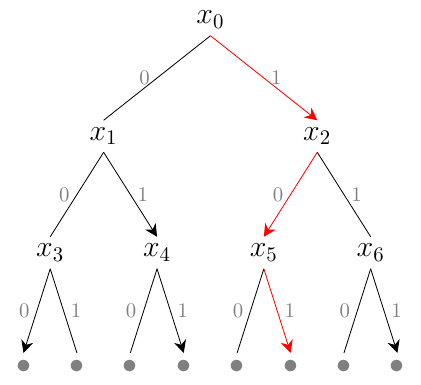}
        \vspace*{0.4em}
        \caption{
            Every function $f: ~ \cX \to \zo$ defines a \emph{path} in the tree, which is a sequence $u_0,u_1,u_2,\dots,u_{d-1}$, where $u_0$ is the root, $d$ is the depth of the tree, and for each $i \in [d-1]$, $u_{i}$ is the $b$-child of $u_{i-1}$ with $b = f(u_{i-1}) \in \zo$. This figure shows that the function $f$ from \cref{subfigure:tree-function} has $\treepath{f} = \paren*{x_0,x_2,x_5}$, depicted in {\color{red}red}. In particular, $x_2$ is `on-path' for $f$, but $x_6$ is `off-path' for $f$.
        }
        \label{subfigure:tree-path}
    \end{subfigure}
    \hfill
    \begin{subfigure}[T]{0.48\textwidth}
        \centering
        \includegraphicsorstandalone[width=\largetreewidth\linewidth]{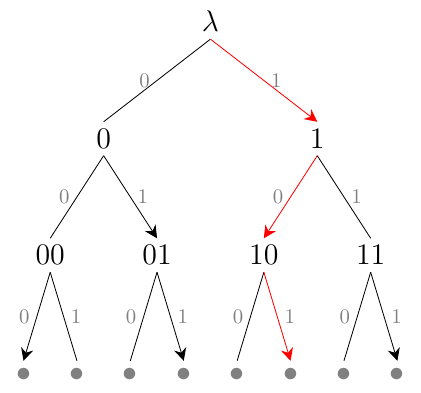}
        \caption{In this paper we use a naming convention where, without loss of generality, we identify the domain elements $x_i$ that are assigned to nodes with bit strings. The root is identified with the empty string $\lambda$, and for each pair of nodes $u$, $v$ such that $u$ is the $b$-child of $v$ (for $b \in \zo$), we have $u = v \circ b$, where `$\circ$' denotes string concatenation. (Because the $x_i$'s may not be distinct, a domain element may be identified with more than one bit string.)}
        \label{subfigure:tree-naming-convention}
    \end{subfigure}
    \caption{Paths in trees.}
    \label{figure:trees}
\end{figure}

\section{Technical Overview}

In this section we explain some of the main ideas in our proofs. Formal definition appear in \cref{section:preliminaries}. Full formal statements of the results, as well as detailed rigorous proofs, appear in \cref{section:lower-bound,section:sequence-length,section:upper-bound}.

\subsection{Paths in Trees}

We make extensive use of the following notion. Given a perfect binary tree $\binarytree{d}$ of depth $d$, every function $f: ~ \binarytree{d} \to \zo$ defines a unique \emph{path} in the tree. The path is a sequence of nodes $\treepath{f} = \paren*{x_{i_0},x_{i_1},\dots,x_{i_d}}$, as explained in \cref{subfigure:tree-path}. See \cref{section:preliminaries} for formal definitions.

\subsection{Proof Ideas for the Lower Bound}
\label{section:overview:lower-bound}

We start with an elementary observation about the adversary's dilemma in the transductive online learning setting. Before round $t$ of the game, the adversary selected a full sequence of instances $x_1,x_2,\dots,x_n \in \cX$, and assigned some initial labels $y_1,y_2,\dots,y_{t-1} \in \zo$. At the start of round $t$, the adversary must consider the \emph{version space},
\[
    \cH_t = \big\{h \in \cH: ~ \paren*{\forall i \in [t-1]: ~ h(x_i) = y_i}\big\}.
\]
If all $h \in \cH_t$ assign $h(x_t) = b$ for some $b \in \zo$, then the adversary has no choice but to assign the label $y_t = b$. Otherwise, the adversary can \emph{force a mistake} at time $t$. Namely, after seeing the learner's prediction $\hat{y}_t$, the adversary can assign $y_t = 1-\hat{y}_t$, incrementing the number of learner mistakes by $1$. 

But ``just because you can, doesn't mean you should''. If the adversary is greedy and forces a mistake at time $t$, they may pay dearly for that later. As an extreme example, consider the case where there is a single $h_1 \in \cH_t$ that assigns $h_1(x_t) = 1$, and all other functions $h \in \cH_t$ assign $h(x_t) = 0$. If the learner selects $\hat{y}_t = 1$ and the adversary forces a mistake at time $t$, the version space at all subsequent times $s > t$ will be $\cH_s = \set*{h_1}$, and the adversary will be prevented from forcing any further mistakes.

A natural strategy for the adversary is therefore to be greedy up to a certain limit. Namely, at each time $t$ the adversary computes the ratio\footnote{%
    For a class $\cH$ of Littlestone dimension $d$, the adversary will use only a subset of $\cH$ of cardinality $2^{d}$ that shatters a Littlestone tree of depth $d-1$. So without loss of generality, we assume that $\cH$ has cardinality $2^d$ (in particular, $\cH$ is finite), and the ratio is well-defined.%
}
\begin{equation*}
    r_t = \frac{|\{h \in \cH_t: ~ h(x_t) = 1\}|}{|\cH_t|}.
\end{equation*}
If $r_t \in [\varepsilon,1-\varepsilon]$ for some parameter $\varepsilon > 0$ (``the version space is not too unbalanced''), then the adversary forces a mistake. Otherwise, the adversary assigns the majority label, i.e., $y_t = \indicator{r_t \geq 1/2}$. This ensures that the version space does not shrink too fast:
\begin{itemize}
    \item{
        If no mistake is forced, then $|\cH_{t+1}| \geq (1-\varepsilon)\cdot|\cH_t|$, and
    }
    \item{
        If a mistake is forced, $|\cH_{t+1}| \geq \varepsilon\cdot|\cH_t|$.
    }
\end{itemize}
In particular, at the end of the game, the version space $\cH_{n+1}$ is of size
\begin{equation}
    \label{eq:version-space-lower-bound}
    \abs*{\cH_{n+1}} \geq \varepsilon^M \cdot (1-\varepsilon)^{n-M} \cdot \abs*{\cH} \geq \varepsilon^M \cdot (1-\varepsilon)^{n} \cdot 2^d,
\end{equation}
where $M$ is the number of mistakes that the adversary forces and $n$ is the length of the sequence. The class has size $|\cH| \geq 2^{d}$ because $\LD{\cH} = d$, and by removing functions from the class if necessary (which can only make learning easier), we may assume without loss of generality that $|\cH| = 2^{d}$. Namely, the class precisely shatters a Littlestone tree of depth $d-1$ such that for every assignment of labels to a root-to-leaf path in the tree, the class contains exactly one function that agrees with that assignment (see \cref{definition:littlestone-tree-and-dimension} for detail).

Notice that we have not yet specified how the adversary selects the sequence $x$. While the adversary's labeling strategy is extremely simple (determined by the ratio $r_t$ and the prediction $\hat{y}_t$), constructing of the sequence $x$ requires some care, to ensure that it has the following two properties:
\begin{itemize}
    \item{
        \textbf{Property I:} 
        The length $n$ of the sequence satisfies $n = 2^{\Thetasqrtd}$, and %
    }
    \item{
        \textbf{Property II:} 
        For every sequence of predictions $\hat{y}_1,\dots,\hat{y}_n$ selected by the learner, the resulting sequence of labels $y_1,\dots,y_n$ selected by the adversary are consistent with some function $h \in \cH$ such that $x$ contains all the nodes in $\treepath{h}$.\footnote{%
            Recall that the \emph{path} of a function $h$ is depicted in \cref{subfigure:tree-path}, and defined in \cref{definition:path-in-tree}.%
        }
    }
\end{itemize}
These properties can be achieved by carefully simulating all possible execution paths of the adversary. 

Observe that if $\treepath{h} = \paren*{u_1,\dots,u_d}$ then the sequence of labels $h(u_1),h(u_2),\dots,h(u_{d})$ uniquely identifies the function $h$ within the class $\cH$. Hence, Property II and the assumption $|\cH|=2^d$ imply that at the end of the game, the version space $\cH_{n+1}$ has cardinality 
\begin{equation}
    \label{eq:H-is-singleton}
    \abs*{\cH_{n+1}} = 1.
\end{equation}
Combining Property I ($n = 2^{\Thetasqrtd}$), \cref{eq:H-is-singleton,eq:version-space-lower-bound}, and choosing $\varepsilon=2^{-\Thetasqrtd}$ gives
\[
    1 \geq \varepsilon^M \cdot (1-\varepsilon)^n \cdot 2^d \geq 2^{-\ThetaOf{M \cdot \sqrt{d}}} \cdot 2^d,
\]
\ifneurips{which implies $M = \Omegasqrtd$, as desired.}\else{which implies
\[
    M = \Omegasqrtd,
\]
as desired.}\fi

\subsection{Proof Ideas for the Upper Bound}
\label{section:overview:upper-bound}

In this section we explain the main ideas in the proof of \cref{theorem:separation}, which states that for every $d \in \bbN$, there exists a class of Littlestone dimension $d$ that is learnable in the transductive online setting with a mistake bound of $\Osqrtd$.

Of course, not every Littlestone class satisfies this property. For instance, the set of all functions $[d] \to \zo$ has Littlestone dimension $d$, but the adversary can force the learner to make $d$ mistakes when learning this class in the transductive setting.\footnote{%
    The adversary simply selects the sequence $x = \paren{1,2,3,\dots,d}$, and for each $x_i$, the adversary forces a mistake by selecting $y_i = 1-\hat{y}_i$. The adversary's choice of labels is realizable because we are working with the class of all function $[d] \to \zo$.%
} So our task in this proof is to construct a class that is especially easy to learn in the transductive setting (i.e., learnable with $\Osqrtd$ mistakes), while still being hard (requiring $d$ mistakes) in the standard setting.

\subsubsection{Sparse Encodings are Easy to Guess}
\label{section:sparse-encoding}

We start with an elementary observation. Consider the following two bit strings:
\begin{align*}
    \text{Binary:} &~  110101
    \\
    \text{One-hot:} &~ 0000000000000000000000000000000000000000000000000000100000000000
\end{align*}
Both of these strings encode the number $53$. However, one of the encodings is much easier to guess than the other: suppose we are tasked with guessing the bits in an encoding of an integer between $0$ and $2^6-1$. We guess the bits one at a time, and after each guess, an adaptive adversary tells us whether our guess was correct.

Now, if the bit string is a binary encoding, the task is hard. Each bit can either be $0$ or $1$, regardless of the values of the previous bits, and so the adversary can force a mistake on every bit. On the other hand, if we know that the string is a one-hot encoding, there exists an attractive strategy --- always guess $0$. This ensures that we will make at most $1$ mistake. 

Note that at the end of the guessing game we have learned the same amount of \emph{information} (for a number between $0$ and $2^n-1$, we learned $n$ bits of information), but the number of \emph{mistakes} is very different ($n$ mistakes vs.\ $1$ mistake).

\subsubsection{Construction of the Hypothesis Class}
\label{section:overview:class}

We now describe a construction of a hypothesis class that is easy to learn in the transductive setting, using the idea of a sparse encoding. Recall that a class $\cH$ has Littlestone dimension at least $d$ (\cref{definition:littlestone-tree-and-dimension} in \cref{section:preliminaries}) if there exists a Littlestone tree of depth $d-1$ such that for every $b \in \zo^d$ there exists $h = h_b \in \cH$ such that the values on the path of $h$ agree with $b$. More formally, $\forall i \in [d]: ~ h(b_{<i}) = b_i$, and in particular $\treepath{h} = \paren*{\lambda, b_{\leq1}, b_{\leq2},b_{\leq3},\dots,b_{\leq d-1}}$. Thus, when constructing a class that shatters a specific Littlestone tree of depth $d-1$, we need to define $2^d$ functions $\set*{h_b: ~ b \in \zo^d}$. For each function $h_b$, the on-path values of the function are fixed (fully determined by $b$), while for the remaining values there is complete freedom (for the nodes $u$ that are off-path we may assign any values $h_b(u) \in \zo$).

Perhaps the simplest way to construct a class of Littlestone dimension $d$ is simply to assign all on-path values as required, and assign $0$ to all other values. Namely, if $u$ is a prefix of $b$ then $h_b(u) = b_{|u|+1}$, and otherwise $h_b(u) = 0$. In a sense, this is the `minimal' class of Littlestone dimension $d$ for a specific Littlestone tree.\footnote{%
    More formally, this is a class with a minimal number of nodes labeled $1$.%
}

Observe that the `minimal' class does not have the desired property of being easy to learn in the transductive setting.\footnote{%
    The adversary can declare a sequence $x$ consisting of all the nodes in the tree in breadth-first order, and then force $d$ mistakes --- one mistake in each layer (depth) of the tree. Specifically, regardless of how the adversary selects the labels, for each $i \in [d]$ there exists a node $u_i$ at depth $i$ that is on-path. When it is time for the learner to predict a label for this $u_i$, the learner knows that $u_i$ is on-path because it has seen the correct labels for all the ancestors of $u_i$. However, the adversary has the freedom to extend the path arbitrarily to the left or to the right, and can therefore force a mistake on $u_i$.%
} However, a certain variation of the `minimal' class that embeds a sparse encoding does satisfy the requirement. In this variation, on-path value of the function $h_b$ are assigned as they must (as determined by $b$), while the off-path values are sampled independently using a biased coin, such that each of them is $0$ with high probability, but has a small probability of being $1$. The probability is chosen carefully so that the class satisfies some simple combinatorial properties, as described further in \cref{section:transition-to-halving,lemma:hypothesis-class}.

\subsubsection{Na\"ive Learning Strategy}

We now explain in broad strokes how the probabilistic construction of the hypothesis class in \cref{section:overview:class} is useful for learning with few mistakes in the transductive setting. 

Notice that when predicting labels for the `minimal' class with nodes in breadth-first order, the learner knows at each step whether they are labeling an on-path or off-path node, because the learner has already seen the correct labels for all ancestors of the current node. For off-path nodes, the learner knows that the true label is $0$, so it never makes mistakes on off-path nodes, but it also gains no new information when the true labels for off-path nodes are revealed. No risk, but no reward either. Instead, all the information about the true labeling function is revealed only at on-path nodes, where the adversary has complete freedom to assign labels and force mistakes. That's why the adversary can force $d$ mistakes.

For the randomly-chosen class, when predicting labels for off-path nodes, the learner may still safely predict a label of $0$. But the reasoning for this is quite different. Conceptually, every off-path label is part of a sparse codeword that identifies the correct labeling function.\footnote{%
    The coin-flips for off-path labels are all independent. For example, if $X$ is a set of nodes all of which are off-path for a subset $H$ of the hypothesis class, then the random variables $\set{h(x): ~ h \in H, ~ x \in X}$ are i.i.d.%
} Because the coin is biased, each bit of the codeword is easy to guess (it is likely to be $0$), but every time that the adversary reveals that the true label for an off-path node is indeed $0$, the learner gains a small (nonzero) amount of information about the true labeling function. Additionally, when the adversary selects an off-path label of $1$, that reveals a lot of information about the true labeling function (such labels are rare in the hypothesis class), and therefore the adversary cannot force many off-path mistakes. Overall, the information about the true labeling function is `smeared' throughout all labels of the tree ($0$s and $1$s, on-path and off-path).\footnote{
    Furthermore, the labels for most not-too-small subsets of the nodes reveal a lot of information about the correct labeling function --- not just for a particular subset of nodes.
    These properties led us to code-name this construction while working on the paper as `everything everywhere all at once' (in reference to a 2022 film of that name).
    This is in contrast to the `minimal' function, where the information is concentrated entirely on the function path.
    The asymmetry between the `minimal' class and the probabilistic class is similar to that between the binary and one-hot encodings in \cref{section:sparse-encoding} above.%
}

Thus, the na\"ive general strategy for the learner when using the probabilistically-constructed class is to learn most of the information about the true labeling function by observing off-path labels. By the time the learner reaches an on-path node, it hopefully has already learned enough about the true labeling function in order to make a good prediction on that node.

However, making this general strategy work requires overcoming some very substantial obstacles: 
\begin{enumerate}
    \item{
        Recall that in the transductive setting, the adversary can present the nodes of the tree in any order of its choosing --- it does not have to present the tree in breadth-first order. The na\"ive strategy works only if the learner sees many off-path nodes before it sees most on-path nodes. But what happens if the adversary decides to present many on-path nodes near the beginning of the sequence? To handle this, the learner incorporates a strategy we call `danger zone minimization', as described in \cref{section:danger-zone-minimization}. 
    }
    \item{
        Another, equally problematic, issue also arises from the fact that the sequence presented by the adversary might not be in breadth-first order. Recall that breadth-first order\footnote{As well as depth-first order.} has the property that for every node $u$ in the sequence, all the ancestors of $u$ appear \emph{before} $u$ in the sequence. This means that by the time the learner needs to predict a label for $u$, the learner knows whether $u$ is on-path or off-path for the true labeling function. But what happens if the adversary presents $u$ before some of $u$'s ancestors? Or omits some of $u$'s ancestors from the sequence altogether? In this case the learner doesn't know if $u$ is on-path or off-path, and this presents a double hazard. One hazard is that the leaner doesn't know what label to predict for $u$ --- if $u$ is off-path, the learner can simply predict $0$, but if it is on-path it must do something more elaborate. The second hazard is that, after seeing the correct label for $u$, it is not clear what the learner can infer from it. If $u$ is off-path, its label should be interpreted as part of a sparse encoding of the labeling function. But if $u$ is on-path, the interpretation must be entirely different. To overcome this challenge, the learner incorporates a strategy we call `splitting experts', described in \cref{section:splitting-experts}.
    }
    \item{
        \label{item:handling-off-path-nodes}
        Limiting off-path mistakes. Thanks to the coin's bias, most off-path nodes have a true label of $0$. Nonetheless, each function in the hypothesis class still has an expected number of $2^{\OmegaOf{d}}$ off-path nodes labeled~$1$, so the learner can afford to misclassify only a vanishing fraction of them! To limit the number of mistakes, the learner extracts information from the sparse encoding and executes a `transition to Halving' strategy, as described in \cref{section:transition-to-halving}.
    }
\end{enumerate}

\subsubsection{Danger Zone Minimization}
\label{section:danger-zone-minimization}

Utilizing information from the `sparse encoding' of the off-path nodes to make good predictions for on-path nodes requires that the learner first see the true labels for many off-path nodes. Until that happens, the learner expects to make many mistakes on on-path nodes. However, whether a node is on-path or off-path is not fixed in advanced --- the adversary may decide this adaptively, in response to the learners predictions. 

\emph{Danger zone minimization} is a strategy used by the learner, to force the adversary to assign few nodes in the beginning of the sequence as on-path (otherwise, if initial nodes are assigned to be on-path by the adversary, then the learner will make few mistakes on those nodes). This is analogous to the standard Halving algorithm (\cref{algorithm:halving}), but instead of minimizing the cardinality of the set of consistent hypotheses (the `version space'), the learner minimizes a subset of the domain (the `danger zone'). 

Concretely, at the beginning of the game the learner initializes a set $S = \set*{x_1,x_2,\dots,x_{\tmax}}$ consisting of the first $\tmax = \tmaxasymptotic$ instances in the sequence $x$ selected by the adversary. This set represents the `danger zone' --- nodes in the beginning of the sequence that have not been labeled yet, that \emph{might} be on-path, and that are not ancestors of a previously-labeled on-path node.\footnote{%
    If $u$ is an ancestor of some on-path node $v$, and $v$ is a $b$-descendant of $u$ for $b\in \zo$, then the true label for $u$ must be $b$.%
} To predict a label for an instance $x_i$, the learner selects a label $\hat{y}_i$ such that if $\hat{y}_i$ is wrong, the danger zone will shrink by at least $1/3$. That is, for $b \in \zo$, if the set $S_b$ of $b$-descendants of $x_i$ has cardinality $|S_b| \geq |S|/3$, the learner predicts $\hat{y}_i = b$. Then, if the adversary selects $y_i = 1-b$, that implies that all $b$-descendants of $x_i$ are off-path for the true labeling functions. Therefore, the learner removes all $b$-descendants of $x_i$ from the danger zone, and the new cardinality is $|S\setminus S_b| \leq (2/3)\cdot|S|$. This guarantees that the learner can make at most $\BigO{\log \tmax} = \Osqrtd$ such mistakes before the danger zone is empty.\footnote{
    Once the danger zone is empty, the learner cannot make any further on-path mistakes within the prefix $x_1,x_2,\dots,x_{\tmax}$. And it will make at most $\Osqrtd$ mistakes on the remaining nodes $x_{\tmax+1},x_{\tmax+2},\dots$, as explained in \cref{section:transition-to-halving}.
}

If neither $S_0$ nor $S_1$ have cardinality at least $|S|/3$, the learner predicts $\hat{y}_i = 0$. If $y_i = 1$ and $x_i$ is on-path for the true labeling function, then the learner updates the danger zone to be $S_0 \cup S_1$,\footnote{Because on-path nodes must be either be descendants or ancestors of $x_i$, and the definition of the danger zone does not require that it contain ancestors of nodes that have been labeled.} again shrinking the danger zone by a factor of at most $2/3$. Otherwise, if $y_i = 1$ and $x_i$ is off-path, then it was an off-path node labeled $1$ (which is rare), and the learner can afford to misclassify it (see \cref{section:transition-to-halving}).

\subsubsection{Splitting Experts}
\label{section:splitting-experts}

The danger zone minimization strategy requires that the learner know whether the node $u$ being classified is on-path or off-path for the true labeling function. However, if $u$ appears in the sequence before some of its ancestors, the learner does not know this. To overcome this difficulty, the learner implements a variant of the standard \emph{multiplicative weights algorithm} using \emph{splitting experts}. This means that initially there is a single expert executing danger zone minimization. When a node $u$ is reached for which danger zone minimization requires knowing whether $u$ is on-path or off-path and that information is not yet evident, each expert is split into two experts, one of which continues the execution of danger zone minimization under the assumption that $u$ is on-path, and the other under the opposite assumption. Thus, at each point in time, there exists precisely one expert for which all path-related assumptions are correct, and therefore that expert will make at most $\Osqrtd$ mistakes. The multiplicative weights algorithm guarantees that the overall number of mistakes will be linear in the number of mistakes of the best expert, i.e.,~$\Osqrtd$.

\subsubsection{Transition to Halving}
\label{section:transition-to-halving}

The hypothesis class is engineered such that it satisfies the following property: there are at most $2^{\BigO{\sqrt{d}}}$ functions in the hypothesis class that agree with any set of $\tmax = \tmaxasymptotic$ labels, or that agree that a set of $\Thetasqrtd$ nodes are all off-path and labeled $1$ (this follows from \cref{lemma:hypothesis-class}). 

Therefore, once the true labels for the first $\tmax$ instances $x_1,x_2,\dots,x_{\tmax}$ have been revealed, or once $\Thetasqrtd$ off-path labels of $1$ have been revealed (whichever happens first), the learner can \emph{transition to halving}: stop doing danger zone minimization, and instead predict the labels for the remaining nodes using the standard Halving algorithm (\cref{algorithm:halving}) on the subset of the hypothesis class that survived. Halving on $2^{\BigO{\sqrt{d}}}$ functions is guaranteed to make at most $\Osqrtd$ mistakes (\cref{claim:halving}).

However, seeing as the learner lacks information on which nodes are off-path, it uses experts, and each expert maintains different path-related assumptions. Thus, each expert decides separately at which point to transition to Halving. The unique expert that makes only correct assumptions will transition `at the right time'. That expert will make at most $\Osqrtd$ mistakes during danger zone minimization, and then at most $\Osqrtd$ additional mistakes during halving.

\subsection{Some Intuition for the Quantity \texorpdfstring{$\sqrt{d}$}{Root d}}

We briefly sketch where the quantity $\sqrt{d}$ arises from. This is a back-of-the-envelope calculation without proof, intended purely as an aid for intuition. Suppose we assigned off-path labels of $1$ with probability $2^{-k}$ instead of $2^{-\sqrt{d}}$. Consider a sequence $x_1,\dots,x_{n}$ of $n = d/2k$ leaves. For any sequence of labels $y_1,\dots,y_n \in \zo$, taking $s = \sum_{i \in [n]}y_i$, there exist roughly  
\[
    2^{d}\cdot\paren*{2^{-k}}^{s}\cdot\paren*{1-2^{-k}}^{n-s} \geq 2^{d}\cdot\paren*{2^{-k}}^n \gg 0
\]
functions in the class for which these leaves are off-path and which agree with the labels $y_1,\dots,y_n$. Therefore, the adversary can force at least $n = \OmegaOf{d/k}$ mistakes.

Similarly, for the sequence $x_1,\dots,x_n$ consisting of all the nodes in the tree of depth at most $k/2$ in breadth-first order, the adversary can force a mistake on every on-path node while assigning a label of $0$ to all off-path nodes, for a total of $k/2$ mistakes. This is true because for any assignment of on-path labels, the fraction of functions which agree with the on-path labels that assign a label of $0$ to all off-path nodes is roughly \ifneurips{$\paren*{1-2^{-k}}^{2^{k/2}} \approx 1$,\nobreakspace}\else
\[
    \paren*{1-2^{-k}}^{2^{k/2}} \approx 1,
\]
\fi{}so in particular for any labeling of the on-path nodes there exists a function in the class that agrees with that labeling and assigns $0$ to all off-path nodes.

Therefore, for any $k$, we obtain a \emph{lower bound} of $\OmegaOf{\frac{d}{k} + k}$ on the number of mistakes. For any $k$, $\frac{d}{k} + k \geq \sqrt{d}$, giving a lower bound of $\Omegasqrtd$. Choosing $k = \sqrt{d}$ to minimize the lower bound will in fact yield a matching upper bound of $\Osqrtd$, as we show in this paper. This completes our overview of the upper bound.

\section{Directions for Future Work}

Following are some interesting open questions:
\begin{enumerate}
    \item{
        Does there exist an efficient learning algorithm that achieves the $\Osqrtd$ upper bound of \cref{theorem:separation}? One needs to be careful about the definition of efficiency here, but one possible formalization is as follows. Does there exist a learning algorithm $A$ and a sequence of classes $\cH_1,\cH_2,\dots$, such that for every $d \in \bbN$:
        \begin{itemize}
            \item{
                $\LD{\cH_d} = d$, and
            }
            \item{
                Given as input the index $d$ and a sequence $x_1,\dots,x_n$, the algorithm $A$ runs in time $\poly{d,n}$ and makes at most $\Osqrtd$ mistakes assuming the labels are realizable by~$\mathcal{H}_d$.
            }
        \end{itemize}
    }
    \item{
         Is there a tradeoff between the cardinality of the domain $\cX$ and the upper bound on the number of mistakes? We used a domain of size roughly $2^d$ in order to obtain our upper bound of $\Osqrtd$. Is it possible to get the same bound with a domain of size $\poly{d}$?
    }
    \item{
        Obtaining more precise asymptotics; for example, is there (an explicit) constant $\alpha > 0$ such that the optimal transductive mistake bound is $\bigl(\alpha + o(1)\bigr) \sqrt{d}$?
    }
\end{enumerate}

\section{Organization}
\vspace*{-1em}
\ifneurips{Complete rigorous mathematical details are deferred to the appendices.\nobreakspace}\fi
Formal definitions appear in \cref{section:preliminaries}. Formal statements and proofs for the lower bound and upper bound appear in \cref{section:lower-bound} and \cref{section:upper-bound}, respectively. Optimal sequence length is discussed in \cref{section:sequence-length}.

    \else
        
    \fi

    \ifdraftcompile
        \section{Preliminaries}
\label{section:preliminaries}

\subsection{Basic Notation}

\begin{notation}
    $\bbN = \set*{1,2,3,\dots}$, i.e., $0 \notin \bbN$. $\logf{\cdot}$ and $\lnf{\cdot}$ denote logarithm to base $2$ and $e$, respectively.
\end{notation}

\begin{notation}[Sequences]
    Let $\cX$ be a set and $n,k \in \bbN$. For a sequence $x = (x_1,\dots,x_n) \in \cX^n$, we write $x_{\leq k}$ to denote the subsequence $(x_1,\dots,x_k)$. If $k\leq0$ then $x_{\leq k}$ denotes the empty sequence, which is also denoted by $\lambda = \cX^{0}$. We use the notation $\cX^{\leq n} = \cup_{i = 0}^n \cX^i$.
\end{notation}

\subsection{Standard Online Learning}
\label{section:preliminaries:standard-online}

Let $\cX$ be a set, and let $\cH \subseteq \zo^{\cX}$ be a collection of functions called a \emph{hypothesis class}. A \emph{learner strategy} or simply \emph{learner} for the standard online learning game (\cref{game:standard}) is a function 
\[
    \learner: ~ \bigcup_{i = 0}^{n-1} \: \paren*{\cX \times \zo}^i \times \cX \to \zo,
\] 
where $n \in \bbN$ is the number of rounds in the game. The set of all such learner strategies is denoted~$\learners_n$. 
An \emph{adversary strategy} or simply \emph{adversary} for the standard online learning game is a pair of functions 
\begin{align*}
    \adversary_{\instance}&: ~ \bigcup_{i = 0}^{n-1} \: \paren*{\cX \times \zo \times \zo}^i \to \cX, \text{ and}
    \\
    \adversary_{\labels}&: ~ \bigcup_{i = 1}^{n-1} \: \paren*{\cX \times \zo \times \zo}^i \times \zo \to \zo.
\end{align*}
The set of all such adversary strategies is denoted $\adversaries_n$. 

Semantically, the interpretation of these strategies is that in each round $t \in [n]$ of \cref{game:standard}, the adversary selects an instance 
\[
    x_t = \adversary_{\instance}(x_1,\hat{y}_1,y_1,\dots,x_{t-1},\hat{y}_{t-1},y_{t-1}) \in \cX,
\]
then the learner makes a prediction
\[
    \hat{y}_t = \learner(x_1,y_1,\dots,x_{t-1},y_{t-1},x_t) \in \zo,
\]
and finally, the adversary assigns a label
\[
    y_t = \adversary_{\labels}(x_1,\hat{y}_1,y_1,\dots,x_{t-1},\hat{y}_{t-1},y_{t-1},\hat{y}_t) \in \zo.
\]
The adversary's function $\adversary_{\labels}$ must satisfy \emph{realizability}, meaning that there exists $h \in \cH$ such that 
\[
    \forall t \in [n]: ~ y_t = h(x_t).
\]
The number of mistakes in a game with $n$ rounds and hypothesis class $\cH$ between learner $\learner$ and adversary $\adversary$ is
\[
    \onlinemistakes{\cH,n,\learner,\adversary} = \abs*{\set*{t \in [n]: ~ \hat{y}_t \neq y_t}}.
\]

\subsection{Transductive Online Learning}

Given $\cX$ and $\cH$ as in \cref{section:preliminaries:standard-online}, a learner strategy for the \emph{transductive online learning setting} (\cref{game:transductive}) is a function 
\[
    \learner: ~ \cX^n \times \bigcup_{i = 0}^{n-1} \: \zo^i \to \zo,
\] 
where $n \in \bbN$ is the number of rounds in the game. An adversary strategy consists of a sequence $x \in \cX^n$ and an \emph{adversary labeling strategy}, which is a function
\[
    \adversary: ~ \paren*{\bigcup_{i = 0}^{n-1} \: \zo^{2i}}\times \zo \to \zo.
\]
The sets of all such learner and adversary strategies are denoted $\learners_n$ and $\adversaries_n$ respectively. 

Semantically, the interpretation of these strategies is that at the start of \cref{game:transductive}, the adversary selects the sequence $x$. Then, in each round $t \in [n]$, the learner makes a prediction
\[
    \hat{y}_t = \learner(x,y_1,\dots,y_{t-1}) \in \zo,
\]
and then the adversary assigns a label
\[
    y_t = \adversary(\hat{y}_1,y_1,\dots,\hat{y}_{t-1},y_{t-1},\hat{y}_t) \in \zo.
\]
Exactly as in \cref{section:preliminaries:standard-online}, the adversary's function $\adversary$ must satisfy realizability, namely,
\[
    \exists h \in \cH ~ \forall t \in [n]: ~ y_t = h(x_t),
\]
and the number of mistakes in a game with sequence length $n$ and hypothesis class $\cH$ between learner $\learner$ and adversary $\adversary$ is
\[
    \transductivemistakes{\cH,n,\learner,\adversary} = \abs*{\set*{t \in [n]: ~ \hat{y}_t \neq y_t}}.
\]

\subsection{Mistake Bounds}

In this paper, we study \emph{optimal mistake bounds}, or the \emph{optimal number of mistakes}, which is the value of \cref{game:standard,game:transductive}. For $M \in \set*{\onlinemistakesname,\transductivemistakesname}$, the optimal number of mistakes in a game with hypothesis class $\cH$ and sequence length $n$ is,
\[
    M(\cH,n) = 
    \inf_{\learner \in \learners_n}
    \:
    \sup_{\adversary \in \adversaries_n} 
    \: 
    M(\cH,n,\learner,\adversary).
\]
The optimal number of mistakes for hypothesis class $\cH$ is
\[
    M(\cH) = 
    \sup_{n \in \bbN} 
    \: 
    M(\cH,n).
\]

\begin{remark}
    As is common in learning theory literature, in both \cref{game:standard} and \cref{game:transductive}, we take the sets $\learners_n$ and $\adversaries_n$ to be the sets of all (deterministic) functions. In this paper, we do not consider randomized strategies. By allowing arbitrary functions, we ignore issues relating to computability.
\end{remark}

\subsection{Trees}

\begin{definition}[Notation for binary trees]
    \label{definition:binary-tree}
    Let $d \in \bbN\cup\{0\}$. A \ul{perfect binary tree of depth $d$} is a collection of $2^{d+1}-1$ nodes, which we identify with the collection of binary strings 
    \[
        \binarytree{d} = \big\{\zo^{k}: ~ k \in \{0,1,2,\dots,d\}\big\}.
    \]
    The empty string, denoted $\lambda = \zo^0$, is a member of $\binarytree{d}$ and is called the \ul{root} of the tree. Every string $u \in \zo^d$ is called a \ul{leaf}. The \ul{depth} of a node $u \in \binarytree{d}$, denoted $|u|$, is the length of $u$ as a string, namely, the integer $k$ such that $u \in \zo^k$.
    
    For two nodes $u,v \in \binarytree{d}$, we say that $u$ is a \ul{parent} of $v$, and that $v$ is a \ul{child} of $u$, if $v = u \circ 0$ or $v = u \circ 1$, where $\circ$ denotes string concatenation. More fully, for $b \in \zo$, we say that $v$ is a \ul{$b$-child} of $u$ if $v = u \circ b$.
    
    Recursively, we define that \ul{$u$ is an ancestor of $v$} and that \ul{$v$ is a descendant of $u$}, and write $u \ancestor v$, if one of the following holds:
    \begin{itemize}
        \item{$u = v$, or}
        \item{$\exists w \in \binarytree{d} ~ \exists b \in \zo: ~ (u \ancestor w) ~ \land ~ (w\circ b = v)$.}
    \end{itemize}
    For $b \in \zo$, we say that $v$ is a \ul{$b$-descendant} of $u$, denoted $u \ancestor_b v$, if $v$ is a descendant of the $b$-child of $u$.
\end{definition}

A function $f: ~ \binarytree{d} \to \zo$ specifies a particular root-to-leaf path in the tree $\binarytree{d}$ (see \cref{figure:trees}). The \emph{on-path} nodes for $f$ are the set of $d+1$ nodes on that root-to-leaf path, as in the following definition.

\begin{definition}[Paths in a binary tree]
    \label{definition:path-in-tree}
    Let $d,k \in \bbN$, $k \leq d$. 
    Let $u \in \zo^k$ be a node in $\binarytree{d}$. The \ul{path to $u$} is the unique sequence $\treepath{u} = \left(u_0,u_1,u_2,\dots,u_k\right)$ such that $u_0 = \lambda$ is the root, $u_k = u$, and $u_i$ is a child of $u_{i-1}$ for all $i \in [k]$.

    Let $f : \binarytree{d} \to \zo$ be a function. The \ul{path of $f$} is the unique sequence $\treepath{f} = \left(u_0,u_1,u_2,\dots,u_d\right)$ such that $u_0 = \lambda$ is the root, and for each $i \in [d]$, $u_i = u_{i-1}\circ f(u_{i-1})$. Namely, $u_i$ is the $f(u_{i-1})$-child of~$u_{i-1}$. 

    For a node $v \in \binarytree{d}$ and a function $f : \binarytree{d} \to \zo$, we write $v \in \treepath{f}$ if $\treepath{f} = \left(u_0,\dots,u_d\right)$ and there exists $i \in \set{0,\dots,d}$ such that $u_i = v$. Otherwise, we write $v \notin \treepath{f}$. 
    
    For a node $v \in \binarytree{d}$ and a set of functions $\cF \subseteq \zo^\binarytree{d}$, we write $v \in \treepath{\cF}$ if
    \[
        \forall f \in \cF: ~ v \in \treepath{f}.
    \]
    Otherwise, we write $u \notin \treepath{\cF}$.
\end{definition}

\subsection{Littlestone Dimension}

\begin{definition}[\citealp{DBLP:journals/ml/Littlestone87}]\label{definition:littlestone-tree-and-dimension}
    Let $\cX$ be a set, let $\cH \subseteq \{0,1\}^\cX$, and let $d \in \bbN \cup \{0
    \}$. We say that \ul{$\cH$ shatters the binary tree $\binarytree{d}$} if there exists a mapping $\binarytree{d} \to \cX$ given by $u \mapsto x_u$
    such that for every $u \in \{0,1\}^{d+1}$ there exists $h_u \in \cH$ such that
    \[
        \forall i \in [d+1]: ~ h(x_{u_{\leq i-1}}) = u_{i}.
    \]
    The \ul{Littlestone dimension} of $\cH$, denoted $\LD{\cH}$, is the supremum over all $d \in \bbN$ such that there exists a Littlestone tree of depth $d-1$ that is shattered by $\cH$.
\end{definition}

Note that by defining the Littlestone dimension this way, every class with Littlestone dimension $d \in \bbN$ contains at least $2^d$ functions. 

\begin{figure}[ht]
    \centering
    \includegraphics[width=0.7\textwidth]{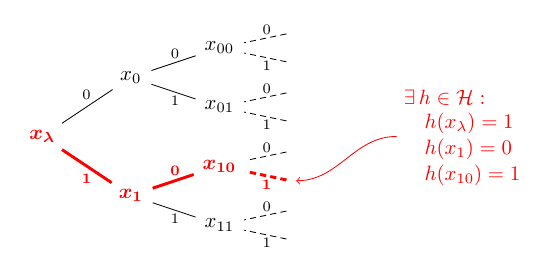}
    \caption{
        A shattered Littlestone tree of depth 2. The empty sequence is denoted by $\lambda$. 
        \\
        \strut\hfill
        {
            \footnotesize
            (Source: \citealp{DBLP:conf/stoc/BousquetHMHY21})
        }
    }
\end{figure}

\begin{theorem}[\citealp{DBLP:journals/ml/Littlestone87}]\label{theorem:ld-is-upper-bound}
    Let $\cX$ be a set and let $\cH \subseteq \{0,1\}^\cX$ such that $d = \LD{\cH} < \infty$. Then there exists a strategy for the learner that guarantees that the learner will make at most $d$ mistakes in the standard (non-transductive) online learning setting, regardless of the adversary's strategy and of the number $n$ of instances to be labeled. Furthermore, there exists an adversary that forces every learner to make at least $\min\,\{n,d\}$ mistakes.
\end{theorem}

    \else
        
    \fi

    \ifdraftcompile
        \section{Lower Bound}
\label{section:lower-bound}

\subsection{Statement}
\label{section:lower-bound-statement}

Our $\Omegasqrtd$ lower bound states the following.

\begin{theorem}[Lower bound]
    \label{theorem:lower-bound}
    There exists a constant $d_0 \geq 0$ as follows. Let $d \in \bbN$, $d \geq d_0$, let $\cX$ be a set, and let $\cH \subseteq \zo^\cX$ be a hypothesis class with $\LD{\cH} = d$. Then there exist a sequence $x \in \cX^{n}$ of length $n = \BigO{d \cdot 2^{\sqrt{d}}}$ and an adversary $\adversary$ that always selects the sequence $x$ and uses a simple adaptive labeling strategy (as in \cref{algorithm:lower-bound-adversary}), such that for every learning rule $\learner$,
    \begin{equation}
        \label{eq:lower-bound-main}
        \transductivemistakes{\cH, n, \learner, \adversary} \geq \sqrt{d}/\constLowerBoundFactor.
    \end{equation}
    Furthermore, for every integer $n \in \bbN$,
    \begin{equation}
        \label{eq:lower-bound-rate}
        \transductivemistakes{\cH,n} \geq \min\,\left\{\!\sqrt{d}/\constLowerBoundFactor, \floor*{\logf{n+1}}\right\}.
    \end{equation}
\end{theorem}

\begin{remark}
    The assumption $\LD{\cH} = d$ implies that for all $k \in [d]$, $\cH$ shatters a Littlestone tree of depth $k$. Thus, the lower bound of \cref{eq:lower-bound-main} in \cref{theorem:lower-bound} immediately implies that for every $k \in [d]$ there exists a sequence $x^{(k)} \in \cX^{n_k}$ of length $n_k = \BigO{k \cdot 2^{\sqrt{k}}}$ such that the adversary $\adversary_k$ that presents the sequence $x^{(k)}$ and assigns labels using the simple labeling strategy of \cref{algorithm:lower-bound-adversary} ensures that for every learner $\learner$, 
    \[
        \transductivemistakes{\cH, n_k, \learner, \adversary_k} \geq \sqrt{k}/\constLowerBoundFactor.
    \]
\end{remark}

See \cref{section:overview:lower-bound} for a general overview of \cref{theorem:lower-bound} and the main proof ideas. 
In the following subsections we prove \cref{theorem:lower-bound}. \cref{algorithm:lower-bound-adversary} gives an explicit construction of the adversary that witnesses the lower bound, using \cref{algorithm:construct-sequence} as a subroutine. We start with presenting some initial observations about the behavior of these algorithms in \cref{section:properties-of-adversary}.

\begin{algorithmFloat}[ht]
    \begin{ShadedBox}
        \textbf{Assumptions:}
        \begin{itemize}
            \item{$d \in \bbN$, $\varepsilon = 2^{-\sqrt{d}/2}$.}
            \item{$T = \binarytree{d}$ is a perfect binary tree of depth $d$.}
            \item{$\cH \subseteq \zo^T$ is a class that shatters $T$.}
        \end{itemize}

        \vspp

        \Adversary\,($\cH$):
        \vsp
        \begin{algorithmic}

            \State $(x_1,x_2,\dots,x_n) \gets \ConstructSequence\,(\cH)$
            \Comment{See \cref{algorithm:construct-sequence}.}

            \vsp

            \State \textbf{send} $(x_1,x_2,\dots,x_n)$ to learner

            \vsp

            \State $\cH_0 \gets \cH$
            \vsp

            \For $t \in [n]$:

                \vsp

                \State \textbf{receive} $\hat{y}_t$ from learner

                \vsp
                
                \State $\displaystyle{r_t \gets 
                \frac{|\{h \in \cH_{t-1}: ~ h(x_t) = 1\}|}{|\cH_{t-1}|}}$

                \vsp
                
                \State $\ymaj \gets \indicator{r_t \geq 1/2}$

                \vsp
                
                \State $y_t \gets \left\{
                    \begin{array}{ll}
                        \ymaj & r_t \notin [\varepsilon, 1-\varepsilon] \\
                        1-\hat{y}_t & \text{otherwise}
                    \end{array}
                    \right.$
                
                \vsp

                \State \textbf{send} $y_t$ to learner

                \vsp

                \State $\cH_t \gets \{h \in \cH_{t-1}: ~ h(x_t) = y_t\}$
            \EndFor
        \end{algorithmic}
    \end{ShadedBox}
    \vspace{-1em}
    \caption{The strategy for the adversary that achieves the lower bound in \cref{theorem:lower-bound}. Note that while the construction of the sequence $x$ is not entirely trivial, the adversary's strategy for labeling this sequence is very simple.}
    \label{algorithm:lower-bound-adversary}
\end{algorithmFloat}

\begin{algorithmFloat}[ht]
    \begin{ShadedBox}
        \textbf{Assumptions:}
        \begin{itemize}
            \item{$d \in \bbN$, $M = \sqrt{d}/\constLowerBoundFactor$, $\varepsilon = 2^{-\sqrt{d}/2}$.}
            \item{$T = \binarytree{d}$ is a perfect binary tree of depth $d$.}
            \item{$\lambda$, the empty string, is the root of $T$.}
            \item{$\cH \subseteq \zo^T$ is a class that shatters $T$.}
        \end{itemize}

        \vsppp

        \ConstructSequence\,($\cH$):
        \vsp
        \begin{algorithmic}

            \State $\cH_{\lambda} \gets \cH$

            \vsp
            
            \State $\bbH_0 \gets \{\cH_{\lambda}\}$
            \Comment{
                A set of classes indexed by bit strings.
            }
            
            \vsp

            \State $\cQ \gets \{\lambda\}$
            \Comment{
                A set of nodes to be processed.
            }

            \vsp

            \State $t \gets 0$

            \vsp

            \While $|\cQ| > 0$:
                
                \vsp

                \State $t \gets t + 1$

                \vsp

                \State{$x_t \gets \text{ arbitrary element from }\cQ$}
                \Comment{\parbox[t]{0.4\linewidth}{Pop an arbitrary element from $\cQ$ and add it to the output sequence.}}
                \vspace*{-1em}
                \State $\cQ \gets \cQ \setminus \{x_t\}$

                \vsp

                \State $\bbH_t \gets \varnothing$

                \vsp

                \For $\cH_b \in \bbH_{t-1}$:
                    \vspp

                    \State $\displaystyle{r 
                        \gets
                        \frac{|\{h \in \cH_b: ~ h(x_t) = 1\}|}{|\cH_b|}}$

                    \vspp

                    \State $\cY \gets
                    \left\{
                    \begin{array}{ll}
                        \zo & \big(r \in [\varepsilon, 1-\varepsilon]\big) \land \big(|b| < M\big) \\
                        \{\indicator{r \geq 1/2}\} & \text{otherwise}
                    \end{array}
                    \right.$
                    \Comment{\parbox[t]{0.23\linewidth}{Adversary will force mistakes on the first $M$ balanced nodes.}}

                    \vsp

                    \For $y \in \cY$:
                        \State $b' \gets
                                \left\{
                                \begin{array}{ll}
                                b  & |\cY| = 1 \\
                                b \circ y & |\cY| = 2
                                \end{array}
                                \right.$
                        \Comment{
                            \parbox[t]{0.42\linewidth}{
                                Restrict class to agree with $y$. If splitting the class in two to force a mistake then create new indices.
                            }
                        }
                        \vspace*{-1em}
                        \State $\cH_{b'} \gets \{h \in \cH_b: h(x_t) = y\}$
                        
                        \vsp

                        \State $\bbH_t \gets \bbH_t \cup \{\cH_{b'}\}$

                        \vsp

                        \If $x_t \in \treepath{\cH_{b'}} \: \land \: |x_t| < d$:
                            \Comment{
                                \parbox[t]{0.44\linewidth}{If $x_t$ is on-path for $\cH_{b'}$ and it has a $y$-child, add that child to $\cQ$.}
                            }
                            \vspace*{-1em}
                            \State $\cQ \gets \cQ \cup \{x_t \circ y\}$
                        \EndIf
                    \EndFor

                \EndFor
                
            \EndWhile

            \vspp

            \State \textbf{return }$\left(x_1,x_2,\dots,x_t\right)$
        \end{algorithmic}
    \end{ShadedBox}
    \vspace{-1em}
    \caption{A subroutine of \cref{algorithm:lower-bound-adversary} for selecting the sequence $x$.}
    \label{algorithm:construct-sequence}
\end{algorithmFloat}

\subsection{Analysis of the Adversary}
\label{section:properties-of-adversary}

\begin{claim}
    \label{claim:sequence-basic-properties}
    Let $d \in \bbN$, let $M = \sqrt{d}/\constLowerBoundFactor$, and let $\cH \subseteq \zo^{\binarytree{d}}$ be a hypothesis class. Consider an execution of $\ConstructSequence\,(\cH)$ as in \cref{algorithm:construct-sequence} that produces a sequence $x_1,x_2,\dots,x_n$. Then:
    \begin{enumerate}[label=(\alph*)]
        \item{
            \label{item:sequence-closed-under-ancestry}
            For all $i \in [n]$, $\treepath{x_i}$ is a subsequence of $x_0,x_1,\dots,x_i$.
        }
        \item{
            \label{item:sequence-is-short}
            The length $n$ of the sequence satisfies 
            $n < \sequencelength{d}$, where $\sequencelength{d} = (d+1)\cdot2^{M+1}$.
        }
    \end{enumerate}
\end{claim}

\begin{proof}~
    \begin{enumerate}[label=(\alph*)]
        \item{
            Fix $i \in [n]$. It suffices to show that for all $u \in \binarytree{d}$, if $u \ancestor x_i$ then $u \in \left(x_1,x_2,\dots,x_i\right)$. Proceed by induction on $i$. For the base case $i=1$, the claim holds because $x_1 = \lambda$. 
            
            For the induction step, assume the claim holds for $i \in [n-1]$. Let $u \ancestor x_{i+1}$, we prove that $u \in (x_1,x_2,\dots,x_{i+1})$. Assume $x_{i+1} \neq \lambda$ (otherwise, there is nothing to prove). 
            
            Because $x_{i+1}$ appears in the sequence $x$, it must have been added to $\cQ$ before it was added to $x$. The only place where items that are not $\lambda$ are added to $\cQ$ is in the line $\cQ \gets \cQ \cup \{x_t \circ y\}$. Namely, there exist an index $j \in [i]$ and a bit $y \in \zo$ such that $x_{i+1} = x_j \circ y$ (note that $j < i+1$ because $x_j$ was added to the sequence before $x_{i+1}$).
            If $x_j = u$ we are done. Otherwise, note that $x_j$ is the parent of $x_{i+1}$, and therefore $u \ancestor x_j$. By the induction hypothesis, $u \in (x_1,x_2,\dots,x_j)$. This concludes the proof.
        }
        \item{
            Items are added to the sequence $x$ only if they were previously added to $\cQ$. By induction on $i \in [n]$, for each $x_i$ in the sequence, there is at most one iteration of the ``while $|Q| > 0$'' loop in which $x_i$ is added to $Q$. The base case $i=1$ holds because $x_1 = \lambda$ is the root, which is added to $Q$ before the while loop, and $\lambda$ is never added to $Q$ within that loop because the line ``$\cQ \gets \cQ \cup \{x_t \circ y\}$'' can only add non-empty bit strings. For the induction step, if the claim holds for all natural numbers $j$ such that $1 \leq j < i \leq n$ then it holds for $i$. Indeed, for $i \geq 2$, $x_i$ can be added to $Q$ only via the line ``$\cQ \gets \cQ \cup \{x_t \circ y\}$'', and only in the iteration of the while loop where $x_t$ is the parent of $x_i$ in the tree $\binarytree{d}$. In that iteration, the parent $x_t$ of $x_i$ is popped from $Q$, which implies that $x_t$ was added to $Q$ in some previous iteration of the while loop ($t < i$), and is no longer in $Q$ after being popped. By the induction hypothesis, $x_t$ will never be added to $Q$ again, and therefore in all subsequent iterations of the while loop $x_t$ will not be the parent of $x_i$, so $x_i$ cannot be added to $Q$ in subsequent iterations via the line ``$\cQ \gets \cQ \cup \{x_t \circ y\}$''.

            Furthermore, if a node $x_i$ is added to $Q$ in some iteration of the while loop, then it remains in $Q$ for the duration of that iteration. So for all $i \in \set{2,3,\dots,n}$, there is precisely one execution of the line ``$\cQ \gets \cQ \cup \{x_t \circ y\}$'' that adds $x_i$ to $Q$. Namely, there is precisely one point in time during the execution of \cref{algorithm:construct-sequence} in which $x_i = x_t \circ y$, $x_i \notin Q$, and the line ``$\cQ \gets \cQ \cup \{x_t \circ y\}$'' is executed resulting in $x_i \in Q$.

            Consider a function $f$ that maps $i \in \set{2,3,\dots,n}$ to the value of the index $b'$ during the unique execution of the line ``$\cQ \gets \cQ \cup \{x_t \circ y\}$'' that adds $x_i$ to $Q$. Namely, if $b'$ had some value $\beta$ when $x_i$ was added to $Q$, then $f(i) = \beta$.
            
            Notice that ``$\cQ \gets \cQ \cup \{x_t \circ y\}$'' is executed only if the condition $x_t \in \treepath{\cH_{b'}}$ is satisfied in the previous line. Furthermore, the line ``$\cH_{b'} \gets \{h \in \cH_b: h(x_t) = y\}$'' ensures that the node $x_i = x_t \circ y$ being added to $Q$ satisfies $x_t \circ y \in \treepath{\cH_{b'}}$, namely
            \[
                \forall h \in \cH_{b'}: ~ x_i \in \treepath{h}.
            \]
            Consequently, $x_i \in \treepath{\cG}$ for any class $\cG$ that is a subset of $\cH_{b'}$; in particular, because the only way that $\cH_{b'}$ might be modified later during the execution of \cref{algorithm:construct-sequence} is by removing elements, it follows that $x_i \in \treepath{\cH_{b'}}$ when the line ``$\cQ \gets \cQ \cup \{x_t \circ y\}$'' is executed and in all subsequent times.
            
            However, $|\!\treepath{\cG}| = d+1$ for any class $\cG \subseteq \zo^{\binarytree{d}}$. This implies that $f$ maps at most $(d+1)$ nodes to each bit string. 
            In other words, for any bit string $b$, the size of the preimage satisfies $|f^{-1}(b)| \leq d+1$.

            The condition ``$|b| < M$'' in \cref{algorithm:construct-sequence} ensures that $|b'| \leq M$, namely, $b' \in \zo^k$ for $k \in \set{0,1,2,\dots,M}$. Thus, 
            \begin{align*}
                n &= 1 + |\{2,3,\dots,n\}|
                \\
                &= 1 + \sum_{\substack{b \in \zo^k \\ k \in \set{0,\dots,M}}} |\{i \in \{2,3,\dots,n\}: ~ f(i) = b\}|
                \\
                &= 1 + \sum_{\substack{b \in \zo^k \\ k \in \set{0,\dots,M}}} |f^{-1}(b)|
                \\
                &\leq 1 + \sum_{\substack{b \in \zo^k \\ k \in \set{0,\dots,M}}}(d+1)
                \\
                &\leq
                1+(d+1)\cdot(2^{M+1}-1).
                \\
                &<
                (d+1)\cdot2^{M+1},
            \end{align*}
            as desired.~\qedhere
        }
    \end{enumerate}
\end{proof}

\vsp

\begin{claim}
    \label{claim:cHt-in-bbHt}
    Let $d \in \bbN$, let $M = \sqrt{d}/\constLowerBoundFactor$, and let $\cH \subseteq \zo^{\binarytree{d}}$ be a hypothesis class. Consider an execution of $\Adversary\,(\cH)$ as in \cref{algorithm:lower-bound-adversary}. Let 
    \[
        \cH_0, \cH_1,\dots,\cH_n
    \]
    be the sequence of hypothesis classes created by $\Adversary$, let
    \[
        S = \big\{t \in [n]: ~ r_t \in [\varepsilon,1-\varepsilon]\big\}
    \]
    be the set of indices where $\Adversary$ forces a mistake, and let
    \[
        \bbH_0,\bbH_1,\dots,\bbH_n
    \]
    be the sequence of collections created by the subroutine $\ConstructSequence$ (\cref{algorithm:construct-sequence}). If $|S| \leq M$ then
    \[
        \forall t \in \{0,1,\dots,n\}: ~ \cH_t \in \bbH_t.
    \]
\end{claim}

\begin{proof}
    Proceed by induction on $t \in \{0,1,\dots,n\}$. The base case $t = 0$ is satisfied, because $\cH_0 = \cH \in \{\cH\} = \bbH_0$. For the induction step, assume that $\cH_{i-1} \in \bbH_{i-1}$ for some $i \in [n]$. We prove that $\cH_{i} \in \bbH_{i}$.

    Let $y_i$ be the label assigned to $x_i$ by $\Adversary$. Then 
    \[
        \cH_i = \{h \in \cH_{i-1}: ~ h(x_i) = y_i\}.
    \]
    Consider the iteration of the while loop in $\ConstructSequence$ that starts with $t \gets i$. By the induction hypothesis, $\cH_{i-1} \in \bbH_{i-1}$. Therefore, in this iteration of the while loop, there will be an iteration of the ``for $\cH_b \in \bbH_{t-1}$'' loop where $\cH_b = \cH_{i-1}$. In that iteration, $y_i \in \cY$ by construction of $y_i$ and $\cY$. Therefore, in the iteration of the ``for $y \in \cY$'' loop in which $y = y_i$,
    \[
        \cH_{b'} = \{h \in \cH_b: h(x_t) = y\} = \{h \in \cH_{i-1}: h(x_i) = y_i\} = \cH_i.
    \] 
    The class $\cH_{b'}$ is then added to $\bbH_i = \bbH_t$ in the line ``$\bbH_t \gets \bbH_t \cup \{\cH_{b'}\}$''. Furthermore, no class is ever removed from $\bbH_t$. So $\cH_i \in \bbH_i$, as desired.
\end{proof}

\vsp

\begin{claim}
    \label{claim:adversary-has-on-path-node-at-every-depth}
    Let $d \in \bbN$, let $M = \sqrt{d}/\constLowerBoundFactor$, and let $\cH \subseteq \zo^{\binarytree{d}}$ be a hypothesis class. Consider an execution of $\Adversary\,(\cH)$ as in \cref{algorithm:lower-bound-adversary} where the adversary constructs a sequence of nodes $x_1,x_2,\dots,x_n \in \binarytree{d}$ and a sequence of classes $\cH_0,\cH_1,\dots,\cH_n \subseteq \zo^{\binarytree{d}}$. Let
    \[
        S = \big\{t \in [n]: ~ r_t \in [\varepsilon,1-\varepsilon]\big\}
    \]
    be the set of indices where $\Adversary$ forces a mistake, and assume that $|S| \leq M$. Then for all $k \in \{0,1,\dots,d\}$ there exists $i \in [n]$ such that
    \begin{enumerate}
        \item{
            \label{item:xi-depth-k}
            $|x_i| = k$, and
        }
        \item{
            \label{item:xi-on-path}
            $x_i \in \treepath{\cH_{i-1}}$,
        }
    \end{enumerate}
\end{claim}

\begin{proof}
    Proceed by induction on $k$. For the base case $k = 0$, notice that $x_1 = \lambda$, $|\lambda| = 0$, and $\lambda \in \treepath{\cH_{-1}}$.
    
    For the induction step, assume the claim holds for some $k \in \{0,1,\dots,d-1\}$, and take $i_k \in [n]$ such that $|x_{i_k}| = k$ and $x_{i_k} \in \treepath{\cH_{i_k-1}}$; we prove that the claim holds for $k+1$ as well.
    
    Consider the iteration of the while loop in $\ConstructSequence$ in which $x_{i_k}$ is added to the sequence (i.e., the iteration starting with  $t \gets i_k$). By \cref{claim:cHt-in-bbHt} and the assumption $|S| \leq M$, $\cH_{i_k-1} \in \bbH_{i_k-1}$. Hence, within this iteration of the while loop, there is an iteration of the ``for $\cH_b \in \bbH_{t-1}$'' loop such that $\cH_b = \cH_{i_k-1}$. By construction, the set $\cY$ always contains the label predicted by the adversary, so $y_{i_k} \in \cY$. Consider the iteration of the ``for $y \in \cY$'' loop such that $y = y_{i_k}$. By the induction hypothesis, $x_i \in \treepath{\cH_{i_k-1}}$, and since $\cH_{b'} \subseteq \cH_b = \cH_{i_k-1}$, it follows that $x_{i_k} \in \treepath{\cH_{b'}}$. Seeing as $|x_{i_k}| < d$, in the last line of this iteration of the ``for $y \in \cY$'' loop, the node $x_{i_{k+1}} := x_{i_k} \circ y_{i_k}$ is added to $\cQ$. This guarantees that $x_{i_{k+1}}$ will eventually be popped from $\cQ$ and added to the sequence returned by $\ConstructSequence$. Once a node has been added to the sequence, it is never removed.

    Notice that $|x_{i_{k+1}}| = |x_{i_k}| + 1 = k+1$, satisfying \cref{item:xi-depth-k}. Therefore, it remains to show \cref{item:xi-on-path}, namely, to show that $x_{i_{k+1}} \in \treepath{\cH_{i_{k+1}-1}}$.

    Indeed, by the induction hypothesis, $x_i \in \treepath{\cH_{i_k-1}}$, and in the iteration of the ``for $y \in \cY$'' discussed above,  $\cH_{b} = \cH_{i_k-1}$, $\cH_{b'} = \cH_{i_k}$, and $\cH_{b'} = \set*{h \in \cH_b: h(x_{i_k}) = y_{i_k}}$. Hence,
    \[
        \forall h \in \cH_{i_k}: ~ x_{i_k} \in \treepath{h} ~ \land ~ h(x_{i_k}) = y_{i_k}.
    \]
    Seeing as $x_{i_{k+1}} = x_{i_k} \circ y_{i_k}$ This implies that 
    \[
        \forall h \in \cH_{i_k}: ~ x_{i_{k+1}} \in \treepath{h}.
    \]
    \cref{item:xi-on-path} follows from the inclusion $\cH_{i_{k+1}-1} \subseteq \cH_{i_k}$.
\end{proof}

\subsection{Proof}

Finally, we complete the proof of the lower bound.

\begin{proof}[Proof of \cref{theorem:lower-bound}]
    Fix $d_0 = 800$ and assume $d \geq d_0$. Seeing as $\LD{\cH} = d$, $\cH$ shatters the tree $\binarytree{d}$. By replacing $\cH$ with a suitable subset of $\cH$ of cardinality $2^{d+1}$, renaming the elements in the domain of $\cH$ to nodes of $\binarytree{d}$, and restricting the domain of each function in $\cH$ to $\binarytree{d}$, assume without loss of generality that $\cH \subseteq \{0,1\}^{\binarytree{d}}$, $|\cH| = 2^{d+1}$, and $\cH$ shatters $\binarytree{d}$.

    Consider the loop ``for $t \in [n]$'' in \cref{algorithm:lower-bound-adversary}, and let 
    \[
        S = \set*{s_1,s_2,\dots,s_m} = \big\{t \in [n]: ~ r_t \in [\varepsilon, 1-\varepsilon]\big\}
    \]
    be the set of indices where the adversary forces a mistake, such that the learner makes at least $m = |S|$ mistakes. Let $M = \sqrt{d}/\constLowerBoundFactor$, and assume for contradiction that $m \leq M$.

    By \cref{claim:adversary-has-on-path-node-at-every-depth}, there exists $t \in [n]$ such that $|x_t| = d$ (i.e., $x_t$ is a leaf in $\binarytree{d}$) and $x_t \in \treepath{\cH_{t-1}}$, namely,
    \[
        \forall h \in \cH_{t-1}: ~ x_t \in \treepath{h}.
    \]
    Seeing as $x_t$ is a leaf,
    \begin{equation}
        \label{eq:equality-of-leaf-path-and-function-path}
        \forall h \in \cH_{t-1}: ~ \treepath{x_t} = \treepath{h}.
    \end{equation}
    By construction,
    \begin{equation*}
        \label{eq:Ht-agrees-with-sequence}
        \cH_t \subseteq \Big\{h \in \cH: ~ \paren*{\forall i \in [t]: ~ h(x_i) = y_i}\Big\},
    \end{equation*}
    and $\cH_{t}$ is not empty. Fix some $h^* \in \cH_t \subseteq \cH_{t-1}$. By \cref{item:sequence-closed-under-ancestry} in \cref{claim:sequence-basic-properties}, $\treepath{x_t} = \treepath{h^*}$ is a subsequence of $x_1,x_2,\dots,x_t$, so
    \[
        \forall h \in \cH_t ~ \forall x \in \treepath{h^*} : ~ h(x) = h^*(x).
    \]
    Seeing as $\cH$ shatters $\binarytree{d}$ and $|\cH| = 2^{k+1}$, if two functions $h,h^* \in \cH$ agree on the labels for all nodes in $\treepath{h^*}$, then $h = h^*$. We conclude that $\cH_t = \set*{h^*}$ and $|\cH_t| = 1$.

    Consider the loop ``for $t \in [n]$'' in \cref{algorithm:lower-bound-adversary}. For each $t \in [n]$, 
    \[
        |\cH_t| \geq 
        \begin{cases}
            \varepsilon \cdot |\cH_{t-1}|         & t \in S \\
            (1-\varepsilon) \cdot |\cH_{t-1}|     & t \notin S.
        \end{cases}
    \]
    Hence, 
    \begin{align}
        \label{eq:lower-bound-main-calculation}
        1
        &= 
        |\cH_t| 
        \nonumber
        \\
        &\geq 
        \varepsilon^m \cdot (1-\varepsilon)^{n-m} \cdot |\cH_0|
        \nonumber
        \\
        &= 
        \varepsilon^m \cdot (1-\varepsilon)^{n-m} \cdot 2^{d+1}
        \nonumber
        \\
        &\geq
        \varepsilon^m \cdot (1-\varepsilon)^n \cdot 2^{d+1}
        \nonumber
        \\
        &\geq
        \varepsilon^m \cdot (1-\varepsilon)^{\sequencelength{d}} \cdot 2^{d+1}
        \tagexplain{%
            by \cref{item:sequence-is-short} in \cref{claim:sequence-basic-properties}.%
        }
        \nonumber
        \\
        &\geq 
        \varepsilon^m \cdot 2^{d} = 2^{-m\sqrt{d}/2 + d},
    \end{align}
    where the final line holds because $\varepsilon = 2^{-\sqrt{d}/2}$, $\sequencelength{d} = (d+1)\cdot2^{\sqrt{d}/\constLowerBoundFactor+1}$, and
    \begin{equation*}
        \label{eq:calculation-of-e-in-lower-bound}
        \left(1-\varepsilon\right)^{\sequencelength{d}} = \left(1-2^{-\sqrt{d}/2}\right)^{(d+1)\cdot2^{\sqrt{d}/\constLowerBoundFactor+1}} \geq \frac{1}{2}
    \end{equation*}
    for our choice of $d \geq 800$. Rearranging \cref{eq:lower-bound-main-calculation} yields
    \[
        2\sqrt{d} \leq m.
    \]
    This is a contradiction to the assumption $m \leq M = \sqrt{d}/\constLowerBoundFactor$. We conclude that an adversary $\adversary$ following \cref{algorithm:lower-bound-adversary} satisfies
    \begin{equation}
        \label{eq:lower-bound-concluded}
        \inf_{\learner \in \learners_n} ~ \transductivemistakes{\cH, n, \learner, \adversary} \geq m > M = \sqrt{d}/\constLowerBoundFactor,
    \end{equation}
    as desired.

    To establish the ``furthermore'' part of the theorem, fix a length $n \in \bbN$. Let $k$ be the largest integer such that $2^{\ceil*{\!\sqrt{k}/\constLowerBoundFactor}} \leq n+1$ and $k \leq d$. By \cref{eq:lower-bound-concluded}, there exists some sequence on which the adversary can force every learning rule to make at least $\ceil*{\!\sqrt{k}/\constLowerBoundFactor}$ mistakes. By \cref{theorem:minimal-sequence}, this implies that there exists a sequence of length $2^{\ceil*{\!\sqrt{k}/\constLowerBoundFactor}} - 1 \leq n$ on which the adversary can force every learning rule to make at least $\ceil*{\!\sqrt{k}/\constLowerBoundFactor} = \min\,\left\{\!\ceil*{\!\sqrt{d}/\constLowerBoundFactor}, \floor*{\logf{n+1}}\right\}$ mistakes. Namely,  
    \[
        \transductivemistakes{\cH,n} \geq \min\,\left\{\!\ceil*{\!\sqrt{d}/\constLowerBoundFactor}, \floor*{\logf{n+1}}\right\},
    \]
    as in \cref{eq:lower-bound-rate}.
\end{proof}

    \else
        
    \fi

    \ifdraftcompile
        \section{Sequence Length}
\label{section:sequence-length}

In this section, we show that if there exists a sequence on which the adversary can force $M$ mistakes, then a sequence of length $2^M-1$ is sufficient, and this upper bound is tight for some classes.\footnote{
    Of course, there also exist classes for which a shorter sequence is sufficient. For instance, if the class shatters (in the VC sense) a subset of the domain of cardinality $M$, then a sequence of length $M$ suffices.
}

\begin{definition}[Minimal sequence]
    Let $\cX$ be a set, let $\cH \subseteq \zo^\cX$ be a class, and let $M \in \bbN$.

    The \ul{minimal sequence length for forcing $M$ mistakes for the class $\cH$}, denoted $\minimalsequence{\cH,M}$ is
    \[
        \minimalsequence{\cH,M} = \inf ~ \set*{%
            n \in \bbN:
            ~
            \paren*{%
                \exists x \in \cX^n:
                ~ 
                \transductivemistakes{\cH,x}
                \geq M
            }
        }.
    \]
    In words, $\minimalsequence{\cH,M}$ is the smallest integer $n$ for which there exists a sequence of length $n$ on which the adversary can force at least $n$ mistakes; if no such sequence exists, then ${\minimalsequence{\cH,M} = \infty}$.
\end{definition}

\begin{theorem}[Minimal sequence bound]
    \label{theorem:minimal-sequence}
    Let $\cX$ be a set, and fix $M \in \bbN$. Then for any class $\cH \subseteq \zo^\cX$, if $\minimalsequence{\cH,M} < \infty$ then 
    \[
        \minimalsequence{\cH,M} \leq 2^{M}-1.
    \] 
    Furthermore, there exists a class $\cH \subseteq \zo^\cX$ for which $\minimalsequence{\cH,M} = 2^{M}-1$.
\end{theorem}

\cref{theorem:minimal-sequence} is a corollary of the tree rank characterization of $\transductivemistakesname$ from \cite{DBLP:journals/ml/Ben-DavidKM97}. For completeness, we present a direct proof of \cref{theorem:minimal-sequence} that does not directly invoke that characterization. Roughly, given an adversary $\adversary_0$ that forces every learner to make at least $M$ mistakes on a (possibly long) sequence $x$, we apply two modifications to obtain new adversaries 
\[
    \adversary_0 \leadsto \adversary_1 \leadsto \adversary_2.
\]
$\adversary_1$ forces $M$ mistakes and has a specific structure that we call `rigidity', but it still uses the same (possibly long) sequence $x$. Capitalizing on the rigid structure, $\adversary_2$ selects a subsequence of $x$ of length at most $2^{M}-1$, and forces $M$ mistakes on that subsequence.

\subsection{Rigid Adversary}

\begin{definition}[Rigid adversary]
    Let $n \in \bbN$, let $\cX$ be a set, and let 
    \[
        \adversary: ~ \paren*{\bigcup_{k = 0}^{n-1}\:\zo^{2k}} \times \zo \to \zo
    \] 
    be an adversary strategy for some fixed sequence $x \in \cX^n$. We say that $\adversary$ is \ul{rigid} if there exists a function 
    \[
        f: ~ \bigcup_{k = 0}^{n-1}\:\zo^{k} \to \set*{0,1,\mistake}
    \]
    such that for all $k \in \set{0,1,\dots,n-1}$ and all $y,\hat{y} \in \zo^k$,
    \[
        \adversary\paren*{
            \hat{y}_1,y_1,\dots,\hat{y}_{k},y_k,\hat{y}_{k+1}
        }
        =
        \left\{
        \begin{array}{ll}
        f(y_1,\dots,y_k) & \qquad f(y_1,\dots,y_k) \in \zo \\
        1-\hat{y}_{k+1} & \qquad f(y_1,\dots,y_k) = \mistake
        \end{array}
        \right..
    \]
\end{definition}

Note that if an adversary is rigid, then the function $f$ that witnesses this is uniquely determined.

\begin{claim}[Rigid adversary exists]
    \label{claim:rigid-strategy}
    Let $n, M \in \bbN$, let $\cX$ be a set, let $x \in \cX^n$, and let $\cH \subseteq \zo^\cX$ be a class. Let $\adversary$ be an adversary strategy that forces every learner to make at least $M$ mistakes on~$x$. Then there exists an adversary strategy $\adversary^*$ such that:
    \begin{enumerate}
        \item{
            \label{item:rigid-strategy-exists}
            $\adversary^*$ forces every learner to make at least $M$ mistakes on $x$ and $\adversary^*$ is rigid.
        }
        \item{
            \label{item:rigid-strategy-m-stars}
            Let $f$ be the function that witnesses the rigidity of $\adversary^*$. Then for every $y \in \zo^n$, the sequence
            \[
                f(y_{\leq 0}), f(y_{\leq 1}), f(y_{\leq 2}), \dots, f(y),
            \]
            has at least $M$ members equal to $\mistake$. 
        }
    \end{enumerate}
\end{claim}

\begin{proof}[Proof of \cref{claim:rigid-strategy}]
    For \cref{item:rigid-strategy-exists}, consider the adversary strategy $\adversary^*$ that simulates an execution of $\adversary$, as in \cref{algorithm:rigid-adversary}. In broad strokes, $\adversary^*$ functions as a middle-man between the learner and $\adversary$. As the learner makes a sequence of predictions $\hat{y} \in \zo^n$, the adversary $\adversary^*$ generates a sequence of (possibly different) predictions $\tilde{y} \in \zo^n$, and sends those to the adversary $\adversary$. Adversary $\adversary$ sees only the predictions $\tilde{y}$, and assigns labels $y \in \zo^n$, which are relayed back to the learner by $\adversary^*$ with no modifications.

    \begin{algorithmFloat}[!ht]
        \begin{ShadedBox}
            \textbf{Assumptions:}
            \begin{itemize}
                \item{$n \in \bbN$, $\cX$ is a set, $x \in \cX^n$ is a fixed sequence of instances.}
                \item{$\adversary: ~ \paren*{\bigcup_{k = 0}^{n-1}\:\zo^{2k}} \times \zo \to \zo$ is an adversary labeling strategy for $x$.}
            \end{itemize}

            \vspp

            \textsc{RigidAdversary}:
            \vsp
            \begin{algorithmic}

                \State \textbf{send} $x_1,\dots,x_n$ to the learner

                \vsp

                \For $t = 1,2,\dots,n$:
                    \vsp
                    \State \textbf{receive} prediction $\hat{y}_t$ from learner

                    \vsp

                    \If $\adversary\paren*{
                            \tilde{y}_1,y_1,\dots,\tilde{y}_{t-1},y_{t-1},0
                        } = 0$:
                        \State $\tilde{y}_t \gets 0$
                    \ElsIf $\adversary\paren*{
                            \tilde{y}_1,y_1,\dots,\tilde{y}_{t-1},y_{t-1},1
                        } = 1$:
                        \State $\tilde{y}_t \gets 1$
                    \Else:
                        \State $\tilde{y}_t \gets \hat{y}_t$
                    \EndIf

                    \vsp

                    \State \textbf{send} prediction $\tilde{y}_t$ to $\adversary$

                    \vsp

                    \State \textbf{receive} label $y_t$ from $\adversary$

                    \vsp

                    \State \textbf{send} label $y_t$  to learner
                \EndFor
            \end{algorithmic}
        \end{ShadedBox}
        \vspace{-1em}
        \caption{Construction of a rigid adversary, by simulating a given adversary $\adversary$.}
        \label{algorithm:rigid-adversary}
    \end{algorithmFloat}
    First, observe that $\adversary^*$ satisfies the realizability requirement. Indeed, $\adversary^*$ simulates an execution of $\adversary$ such that the sequence of labels $y_1,\dots,y_n$ sent by $\adversary^*$ to the learner is exactly the sequence of labels selected by $\adversary$. Seeing as $\adversary$ is realizable, every sequence of labels selected by $\adversary$ is realizable, and therefore every sequence of labels selected by $\adversary^*$ must be realizable as well.

    Second, observe that $\adversary^*$ forces every leaner to make at least $M$ mistakes. To see this, notice that in \cref{algorithm:rigid-adversary},
    \begin{equation}
        \label{eq:y-tilde-at-least-M-mistakes}
        \sum_{t \in [n]} \indicator{\tilde{y}_t \neq y_t} \geq M.
    \end{equation}
    Indeed, $\adversary$ forces every learner to make at least $M$ mistakes, and in particular this applies to a learner that makes predictions $\tilde{y}$ as in the simulation. Furthermore, observe that $\adversary^*$ only alters the predictions it receives from the learner in cases when it selects a label that is accepted by $\adversary$, namely,
    \begin{equation}
        \label{y-tilde-y-hat-equal}
        \forall t \in [n]: ~ \tilde{y}_t \neq \hat{y}_t \implies \tilde{y}_t = y_t.
    \end{equation} 
    Therefore, if $E = \set*{t \in [n]: ~ \tilde{y}_t = \hat{y}_t}$, then
    \begin{align}
        \label{eq:y-hat-more-mistakes-than-y-tilde}
        \sum_{t \in [n]} \indicator{\tilde{y}_t \neq y_t}
        &=
        \sum_{t \in E} \indicator{\tilde{y}_t \neq y_t}
        +
        \sum_{t \in [n]\setminus E} \indicator{\tilde{y}_t \neq y_t}
        \nonumber
        \\
        &=
        \sum_{t \in E} \indicator{\tilde{y}_t \neq y_t} + 0
        \tagexplain{By \cref{y-tilde-y-hat-equal}}
        \nonumber
        \\
        &=
        \sum_{t \in E} \indicator{\hat{y}_t \neq y_t}
        \tagexplain{Defintion of $E$}
        \nonumber
        \\
        &\leq
        \sum_{t \in [n]} \indicator{\hat{y}_t \neq y_t}.
    \end{align}
    Combining \cref{eq:y-tilde-at-least-M-mistakes,eq:y-hat-more-mistakes-than-y-tilde} implies that $\adversary$ forces at least $M$ mistakes.

    Third, we show that $\adversary^*$ is rigid. We claim that there exists a function $g: \zo^{\leq n-1} \to \zo^{\leq n-1}$ such that for every $t \in \set*{0,1,2,\dots,n-1}$,
    \[
        \paren*{\tilde{y}_1,\dots,\tilde{y}_t}
        =
        g(y_1,\dots,y_t).
    \]
    Proceed by induction on $t$. For the base case $t = 0$ there is nothing to prove. For the induction step, we assume the claim holds for some $t = k < n-1$, and show that it holds for $t = k+1$. From \cref{algorithm:rigid-adversary}, $\tilde{y}_{k+1}$ satisfies
    \begin{equation}
        \label{eq:rigid-construction}
        \tilde{y}_{k+1}
        =
        \left\{
        \begin{array}{ll}
        0 & 
        \adversary\paren*{
            \tilde{y}_1,y_1,\dots,\tilde{y}_{k},y_k,0
        } = 0
        \\
        1 &
        \adversary\paren*{
            \tilde{y}_1,y_1,\dots,\tilde{y}_{k},y_k,0
        } = \adversary\paren*{
            \tilde{y}_1,y_1,\dots,\tilde{y}_{k},y_k,1
        } = 1
        \\
        1-y_{k+1} & \text{otherwise}
        \end{array}
        \right..
    \end{equation}
    The first two cases in \cref{eq:rigid-construction} are immediate from \cref{algorithm:rigid-adversary}, and the remaining case occurs when $\adversary$ forces a mistake at time $k+1$, namely, when $\adversary$ selects $y_{k+1} = 1-\tilde{y}_{k+1}$.
    Thus, $\tilde{y}_{k+1}$ is a function of $y_{\leq k+1}$ and $\tilde{y}_{\leq k}$. By the induction hypothesis, $\tilde{y}_{\leq k} = g(y_{\leq k})$, so $\tilde{y}_{k+1}$ is simply a function of $y_{\leq k+1}$. This establishes the existence of the desired function $g$.

    Hence, $\adversary^*$ is rigid, as witnessed by the function
    \[
        f\paren*{
            y_1,\dots,y_k
        }
        =
        \left\{
        \begin{array}{ll}
        0 & 
        \adversary\paren*{
            \tilde{y}_1,y_1,\dots,\tilde{y}_{k},y_k,0
        } = 0
        \\
        1 &
        \adversary\paren*{
            \tilde{y}_1,y_1,\dots,\tilde{y}_{k},y_k,0
        } = \adversary\paren*{
            \tilde{y}_1,y_1,\dots,\tilde{y}_{k},y_k,1
        } = 1
        \\
        \mistake & \text{otherwise}
        \end{array}
        \right.,
    \]
    where $f$ is a well-defined function because $\tilde{y}_{\leq k} = g(y_{\leq k})$.

    We have seen that $\adversary^*$ is a valid (realizable) adversary that forces every learner to make at least $M$ mistakes, and it is rigid. This concludes the proof of \cref{item:rigid-strategy-exists}.

    Finally, For \cref{item:rigid-strategy-m-stars}, note that $\tilde{y}_t \neq y_t$ only if $\adversary$ forces a mistake at time $t$ in the sense that $\adversary$ selects $y_t = 1-b$ for any prediction $b \in \zo$ provided at time $t$. If $\adversary$ forces a mistake at time $t$, then $\adversary^*$ forces a mistake at time $t$ as well. Therefore, if $\tilde{y}_t \neq y_t$, then $f(y_{<t}) = \mistake$, namely, $\tilde{y}_t$ makes mistakes only when the value of $f$ is $\mistake$. By \cref{eq:y-tilde-at-least-M-mistakes}, $\tilde{y}_t$ makes at least $M$ mistakes throughout the game, so there must be at least $M$ rounds where $f$ outputs $\mistake$, as desired.
\end{proof}

\subsection{Essential Indices}

\begin{definition}
    \label{definition:essential-index}
    Let $n, M \in \bbN$, let $\cX$ be a set, let $x \in \cX^n$, and let $\cH \subseteq \zo^\cX$ be a class. Let $\adversary$ be a rigid adversary strategy witnessed by function $f$. We say that an index $t \in [n]$ is \ul{essential for $\adversary$ for forcing $M$ mistakes on $x$} if there exists a sequence $y \in \zo^{t-1}$ such that $f(y) = \mistake$ and the sequence 
    \[
        f(y_{\leq 0}), f(y_{\leq 1}), f(y_{\leq 2}), \dots, f(y_{\leq t-1})
    \]
    contains at most $M - 1$ members equal to $\mistake$.
\end{definition}

\begin{claim}
    \label{claim:few-essential-indices}
    Let $n, M \in \bbN$, let $\cX$ be a set, let $x \in \cX^n$, and let $\cH \subseteq \zo^\cX$ be a class. Let $\adversary$ be a rigid adversary strategy. Then $[n]$ contains at most $2^M-1$ indices that are essential for $\adversary$ for forcing $M$ mistakes on $x$.
\end{claim}

\begin{proof}
    For each essential index $t \in [n]$, there exists a label sequence $y \in \zo^{t-1}$ that witnesses that $t$ is essential, as in \cref{definition:essential-index}. Each  label sequence $y$ is a witness for at most one index (the index $|y|+1$), so it suffices to show that the set $Y \subseteq \zo^{\leq n-1}$ of all witness label sequences is of cardinality at most $2^M-1$. 
    
    Think of $Y$ as a collection of nodes in the binary tree $\binarytree{n-1}$ (\cref{definition:binary-tree}).
    By \cref{definition:essential-index}, if $y \in Y$, then the collection of all ancestors of $y$ in $Y$ has cardinality
    \[
        \big|
            \big\{y_{\leq i}: ~ i \in \set*{0,1,2,\dots,|y|-1}\big\}
            ~ \cap ~ 
            Y
        \big| \leq M-1.
    \]
    Namely, $Y$ is a subtree of depth at most $d = M-1$ in the binary tree $\binarytree{n-1}$.\footnote{%
        The depth of a subtree is $s$ if the longest root-to-node path contains $s+1$ nodes from the subtree.%
    } Hence, the number of nodes in $Y$ is at most
    \[
        2^{d+1}-1 = 2^M-1,
    \]
    as desired.
\end{proof}

\subsection{Proof}

\begin{algorithmFloat}[ht]
    \begin{ShadedBox}
        \textbf{Assumptions:}
        \begin{itemize}
            \vsp
            \item{
                $n,M \in \bbN$, $\cX$ is a set, $x \in \cX^n$ is a fixed sequence of instances.
            }
            \vsp
            \item{
                $\adversary_1: ~ \paren*{\bigcup_{k = 0}^{n-1}\:\zo^{2k}} \times \zo \to \zo$ is a rigid adversary labeling strategy for $x$ that forces every learner to make at least $M$ mistakes on the sequence $x$, and satisfies \cref{item:rigid-strategy-exists,item:rigid-strategy-m-stars} in \cref{claim:rigid-strategy}.
            }
            \vsp
            \item{
                $I = \set*{i_1,i_2,\dots,i_k} \subseteq [n]$ is the set of indices that are essential for $\adversary$ for forcing $M$ mistakes on $x$, and $i_1 \leq i_2 \leq \dots \leq i_k$. By \cref{claim:few-essential-indices}, $k \leq 2^M-1$.
            }
        \end{itemize}

        \vspp

        \textsc{MinimalAdversary}:
        \vsp
        \begin{algorithmic}

            \State \textbf{send} $x_{i_1},x_{i_2},\dots,x_{i_k}$ to the learner

            \vsp

            \For $t = 1,2,\dots,n$:

                \vsp

                \If $t \in I$:
                    \State \textbf{receive} prediction $\hat{y}_t$ from learner
                    
                    \vsp

                    \State \textbf{send} prediction $\hat{y}_t$ to $\adversary_1$
                    
                    \vsp

                    \State \textbf{receive} label $y_t$ from $\adversary_1$
                    
                    \vsp
                    
                    \State \textbf{send} label $y_t$  to learner

                    \vsp

                \Else:

                    \vsp

                    \State \textbf{send} prediction $\hat{y}_t = 0$ to $\adversary_1$

                    \vsp

                    \State \textbf{receive} label $y_t$ from $\adversary_1$

                \EndIf
            \EndFor
        \end{algorithmic}
    \end{ShadedBox}
    \vspace{-1em}
    \caption{Construction of an adversary that forces $M$ mistakes using a sequence $x$ of length at most $2^M-1$. In the proof of \cref{theorem:minimal-sequence}, this adversary is $\adversary_2$. Internally, it simulates a rigid adversary $\adversary_1$.}
    \label{algorithm:minimal-adversary}
\end{algorithmFloat}

\begin{proof}[Proof of \cref{theorem:minimal-sequence}]
    If $\minimalsequence{\cH,M} < \infty$, then there exist a sequence $x \in \cX^n$, and an adversary $\adversary_0$ that forces every learner to make at least $M$ mistakes on $x$. By \cref{claim:rigid-strategy}, there exists a rigid adversary $\adversary_1$ that causes every learner to make at least $M$ mistakes on $x$,\footnote{%
        This is \cref{item:rigid-strategy-exists} in \cref{claim:rigid-strategy}.%
    } and also satisfies \cref{item:rigid-strategy-m-stars} in \cref{claim:rigid-strategy}. Let $f$ be the function that witnesses the rigidity of $\adversary_1$. By \cref{claim:few-essential-indices}, the set $I \subseteq [n]$ of indices that are essential for $\adversary_1$ for forcing $M$ mistakes on $x$ has cardinality $k = |I| \leq 2^M-1$.

    \cref{algorithm:minimal-adversary} defines a new adversary, $\adversary_2$, which forces every learner to make at least $M$ mistakes on a sequence of length $k$. $\adversary_2$ is realizable, because $\adversary_1$ is realizable.\footnote{%
        The argument for realizability is the same as in the proof of \cref{claim:rigid-strategy}.%
    }

    To see that adversary $\adversary_2$ forces every learner to make at least $M$ mistakes, let $y_1,\dots,y_n$ be the sequence of labels assigned by $\adversary_2$. Seeing as $\adversary_2$ assigns the same labels as $\adversary_1$, and $\adversary_1$ satisfies \cref{item:rigid-strategy-m-stars} in \cref{claim:rigid-strategy}, it follows that there are at least $M$ indices $j \in [n]$ such that $f(y_{\leq j-1}) = \mistake$. Fix $J \subseteq [n]$ to be the first $M$ such indices. Then $J \subseteq I$, namely, all the indices in $J$ are essential for $\adversary_1$ for forcing $M$ mistakes on $x$ (\cref{definition:essential-index}). 
    
    Therefore, for each $j \in J$, $\adversary_2$ includes the instance $x_j$ in the sequence of length $k$ sent to the learner. Then, in round $j$ of the $n$ rounds simulated by $\adversary_2$:
    \begin{itemize}
        \item{
            The leaner makes a prediction $\hat{y}_j \in \zo$ corresponding to instance $x_j$.
        }
        \item{
            Adversary $\adversary_2$ sends prediction $\hat{y}_j$ to adversary $\adversary_1$. Because $f(y_{\leq j-1}) = \mistake$, adversary $\adversary_1$ assigns the label $y_j = 1 - \hat{y}_j$. Adversary $\adversary_2$ then sends that label $y_j$ to the learner. So the learner makes a mistake on $x_j$.
        }
    \end{itemize}
    Hence, the learner makes at least $|J| = M$ mistakes, as desired.
\end{proof}

    \else
        
    \fi

    \ifdraftcompile
        \section{Upper Bound}
\label{section:upper-bound}

\subsection{Statement}

The following result states that the lower bound of \cref{theorem:lower-bound} is tight for some classes. 

\begin{theorem}[Upper bound, and separation between standard and transductive online learning]
    \label{theorem:separation}
    For every integer $d \geq \constUpperBoundMinimum$, there exists a hypothesis class $\cH \subseteq \zo^{\cX}$ with a domain $\cX$ of size $|\cX| = 2^d-1$ such that $\LD{\cH} = d$ and the following two conditions hold for all $n \in \bbN$:
    \begin{enumerate}
        \item{
            \label{item:main-theorem-upper-bound} 
            $\transductivemistakes{\cH,n} \leq \constUpperBound\cdot\sqrt{d}$.
        }
        \item{
            \label{item:main-theorem-lower-bound}
            $\onlinemistakes{\cH,n} = \min\,\{n,d\}$.
        }
    \end{enumerate}
\end{theorem}

\subsection{Hypothesis Class}

In this section we construct the hypothesis class for \cref{theorem:separation}.

\begin{lemma}
    \label{lemma:hypothesis-class}
    Let $d \in \bbN$, $d \geq \constClassMinimum$. Let $\binarytree{d}$ be a perfect binary tree of depth $d$, as in \cref{definition:binary-tree}. Then there exists a collection of functions $\cH \subseteq \zo^{\binarytree{d}}$ such that $\LD{\cH} = d+1$ and the following two conditions hold for all $H \subseteq \cH$ and all $X \subseteq \binarytree{d}$:
    \begin{enumerate}
        \item{
            \label{item:class-property-for-0}
            If $\forall h \in H ~ \forall x \in X: ~ x \notin \treepath{h} ~ \land ~ h(x) = 0$, then $\min\,\{|H|,|X|\} < 2^{\constA \sqrt{d}}$.
        }
        \item{
            \label{item:class-property-for-1}
            If $\forall h \in H ~ \forall x \in X: ~ x \notin \treepath{h} ~ \land ~ h(x) = 1$, then $|H| < 2^{\constA \sqrt{d}}$ or $|X| < \constB\sqrt{d}$.
        }
    \end{enumerate}
\end{lemma}

\vsp

The proof employs the probabilistic method, showing that a hypothesis class sampled randomly from a suitable distribution has the desired properties with very high probability.

\begin{proof}
    Let $\cP$ be a probability distribution over hypothesis classes. Formally, $\cP \in \distribution{\paren{\zo^{\binarytree{d}}}^{2^{d+1}}}$ is a distribution over vectors of hypotheses. Each vector $\cH \in \support(\cP)$ consists of $2^{d+1}$ hypotheses,
    \[
        \cH = \paren{h_b}_{b \in \zo^{d+1}},
    \]
    where for each $b \in \zo^{d+1}$, hypothesis $h_b$ is a function $h_b: ~ \binarytree{d} \to \zo$ sampled independently as follows:
    \begin{itemize}
        \item{
            For each $i \in [d] \cup \set{0}$: $h_b(b_{\leq i}) = b_{i+1}$. (In particular, with probability $1$, $\treepath{h_b} = \paren{b_{\leq 0}, b_{\leq 1}, \dots, b_{\leq d}}$, each entry in the vector $\cH$ is unique, and $\cH$ shatters $\binarytree{d}$.)
        }
        \item{
            For each $x \in \binarytree{d} \setminus \treepath{h_b}$, the bit $h_b(x) \in \zo$ is sampled $\Ber{2^{-\sqrt{d}}}$ independently of all other bits in $\cH$, i.e., $\PP{h_b(x) = 1} = \PP{h_b(x) = 1 ~ | ~ \set{h_{b'}}_{b'\neq b},\set{h_b(x')}_{x'\neq x}} = 2^{-\sqrt{d}}$.
        }
    \end{itemize}
    In words, for all nodes on the path in the tree corresponding to $b$, the function $h_b$ assigns a label according to $b$, and for all other nodes, $h_b$ assigns a label of $1$ with probability $2^{-\sqrt{d}}$, and a label of $0$ otherwise. In particular, the collection $\cH$ Littlestone-shatters the tree $\binarytree{d}$.

    \vsp

    Fix $B \subseteq \zo^{d+1}$ and $X \subseteq \binarytree{d}$, and let $E(B,X,y)$ denote the event
    \begin{equation}
        \set*{\forall b \in B ~ \forall x \in X: ~ x \notin \treepath{h_b} ~ \land ~ h_b(x) = y}.
    \end{equation}
    Seeing as each off-path label $h_b(x) \in \zo$ is sampled independently, 
    \begin{align}
        \label{eq:bound-EBXZero}
        \PPP{\cH \sim \cP}{E(B, X, 0)}
        &=
        \prod_{(b,x) \in B \times X} \PPP{\cH \sim \cP}{x \notin \treepath{h_b} ~ \land ~ h_b(x) = 0}
        \nonumber
        \\
        &\leq 
        \paren{1-2^{-\sqrt{d}}}^{|B \times X|}.
    \end{align}
    Hence,
    \begin{flalign}
        \label{eq:bound-exists-EBXZero}
        &
        \PPP{\cH \sim \cP}{\exists B \subseteq \zo^{d+1} ~ \exists ~ X \subseteq \binarytree{d}: ~ E(B,X,0) ~ \land ~ \min\,\{|B|,|X|\} \geq 2^{\constA \sqrt{d}}}
        &
        \nonumber
        \\
        &
        \quad 
        =
        \PPP{\cH \sim \cP}{\exists B \subseteq \zo^{d+1} ~ \exists ~ X \subseteq \binarytree{d}: ~ E(B,X,0) ~ \land ~ |B| = |X| = \ceil*{2^{\constA \sqrt{d}}}}
        &
        \nonumber
        \\
        &
        \quad 
        \leq 
        \binom{|\zo^{d+1}|}{\ceil*{2^{\constA \sqrt{d}}}}\binom{|\binarytree{d}|}{\ceil*{2^{\constA \sqrt{d}}}}\paren{1-2^{-\sqrt{d}}}^{2^{\constATimesTwo \sqrt{d}}}
        \tagexplain{union bound, \cref{eq:bound-EBXZero}}
        &
        \nonumber
        \\
        &
        \quad 
        <
        \binom{2^{d+1}}{2^{\constA \sqrt{d}}+1}^2\cdot\paren{1-2^{-\sqrt{d}}}^{2^{\constATimesTwo \sqrt{d}}}
        \tagexplainmultiline{$2^{\constA \sqrt{d}}$ may not be an integer; formally, we use the generalized binomial coefficient, or simply skip to the next line}
        &
        \nonumber
        \\
        &
        \quad 
        < 
        2^{2\cdot(d+1)\cdot\paren*{2^{\constA \sqrt{d}}+1}}\cdot e^{-2^{-\sqrt{d}} \cdot 2^{\constATimesTwo \sqrt{d}}}
        \tagexplainhspace{$\binom{n}{k} < n^k$ for $k \geq e$; $1+x \leq e^x$ for $x \in \bbR$}{-3em}
        &
        \nonumber
        \\
        &
        \quad 
        < 
        2^{2\cdot(d+2)\cdot2^{\constA \sqrt{d}}}\cdot 2^{-2^{-\sqrt{d}} \cdot 2^{\constATimesTwo \sqrt{d}}}
        \tagexplain{$2^{\constA \sqrt{d}} \geq d+1$ for $d \geq 0$}
        &
        \nonumber
        \\
        &
        \quad 
        = 
        2^{2^{\constA \sqrt{d}}\cdot \paren{2d + 4 - 2^{\constAMinOne \sqrt{d}}}}
        &
        \nonumber
        \\
        &
        \quad 
        < 
        2^{-2^{\constA \sqrt{d}}}.
        \tagexplain{$2d + 4 - 2^{\constAMinOne \sqrt{d}} < -1$ for $d \geq \constClassMinimum$}
        &
    \end{flalign}

    Similarly, 
    \begin{equation}
        \label{eq:bound-EBXOne}
        \PPP{\cH \sim \cP}{\forall b \in B ~ \forall x \in X: ~ x \notin \treepath{h_b} ~ \land ~ h_b(x) = 1}
        \leq 
        2^{-\sqrt{d}\cdot|B \times X|},
    \end{equation}
    so
    \begin{flalign}
        \label{eq:bound-exists-EBXOne}
        &
        \PPP{\cH \sim \cP}{\exists B \subseteq \zo^{d+1} ~ \exists ~ X \subseteq \binarytree{d}: ~ E(B,X,0) ~ \land ~ |H| \geq 2^{\constA \sqrt{d}} ~ \land  ~ |X| \geq \constB\sqrt{d}}
        &
        \nonumber
        \\
        &
        \quad 
        \leq 
        \binom{|\zo^{d+1}|}{\ceil*{2^{\constA \sqrt{d}}}}\binom{|\binarytree{d}|}{\ceil*{\constB\sqrt{d}}}\cdot 2^{-\sqrt{d} \cdot 2^{\constA \sqrt{d}} \cdot \constB\sqrt{d}}
        \tagexplain{union bound, \cref{eq:bound-EBXOne}}
        &
        \nonumber
        \\
        &
        \quad 
        \leq
        \binom{2^{d+1}}{2^{\constA \sqrt{d}}+1} \binom{2^{d+1}}{\constB\sqrt{d}+1} \cdot 2^{-\constB d \cdot 2^{\constA \sqrt{d}}}
        &
        \nonumber
        \\
        &
        \quad 
        <
        2^{(d+1)\cdot\paren*{2^{\constA \sqrt{d}}+1}} \cdot 2^{(d+1)\cdot \paren*{\constB\sqrt{d}+1}} \cdot 2^{-\constB d \cdot 2^{\constA \sqrt{d}}}
        \tagexplain{$\binom{n}{k} < n^k$ for $k \geq e$}
        &
        \nonumber
        \\
        &
        \quad 
        <
        2^{(d+1)\cdot\paren*{2^{\constA \sqrt{d}} + \constB\sqrt{d} + 2}} \cdot 2^{-\constB d \cdot 2^{\constA \sqrt{d}}}
        &
        \nonumber
        \\
        &
        \quad 
        < 
        2^{2d\cdot2^{\constA \sqrt{d}}}\cdot 2^{-\constB d \cdot 2^{\constA \sqrt{d}}}
        \tagexplain{for $d \geq 4$}
        &
        \nonumber
        \\
        &
        \quad 
        < 
        2^{-\constBMinusTwo d 2^{\sqrt{d}}}.
        &
    \end{flalign}

    Applying a union bound to \cref{eq:bound-exists-EBXZero,eq:bound-exists-EBXOne} gives
    \[
        \PPP{\cH \sim \cP}{
            \text{$\cH$ satisfies \cref{item:class-property-for-0,item:class-property-for-1}}
        }
        \geq 1-2^{-2^{\constA \sqrt{d}}}-2^{-\constBMinusTwo d 2^{\sqrt{d}}} \geq 1-10^{-100}.
    \]
    In particular, there exists a collection $\cH$ that satisfies \cref{item:class-property-for-0,item:class-property-for-1}. Furthermore, this collection has $\LD{\cH} = d+1$ (namely, $\LD{\cH} \geq d+1$ because it shatters $\binarytree{d}$; and $\LD{\cH} \leq d+1$ because $|\cH| = 2^{d+1}$).
\end{proof}

\subsection{Algorithm}

In this section we describe \cref{algorithm:upper-bound-main,algorithm:expert-predict,algorithm:expert-extended-update}, which together constitute the learning algorithm that achieves the $\Osqrtd$ mistake upper bound in the transductive setting, as in \cref{theorem:separation}. See \cref{section:overview:upper-bound} for a general overview of these algorithms.

\subsubsection{How Experts Work}

We start with some preliminary remarks about experts in \cref{algorithm:upper-bound-main,algorithm:expert-predict,algorithm:expert-extended-update}.

\paragraph{Experts.}
    A tuple $e = (S,u,H)$ defines an expert that can make predictions using the procedure $\ExpertPrediction(e, \cdot)$. The tuple $e$ reflects two kinds of information:
    \begin{enumerate}
        \item{
            \emph{Knowledge.} Information that the expert \emph{knows} with certainty. Specifically, this reflects the labels $y_1,y_2,\dots$ sent by the adversary so far. All experts see the labels sent by the adversary, so this knowledge is the same for all experts.
        }
        \item{
            \emph{Assumptions.} At certain times, experts make \emph{assumptions} about things that are not known for certain. Specifically, experts assume that certain nodes $x$ are on-path ($x \in \treepath{h}$) or off-path ($x \notin \treepath{h}$) with respect to the correct labeling function $h: ~ \binarytree{d} \to \zo$. Assumptions are simply guesses that may be wrong, and therefore when an expert needs to make such an assumption, it splits into two experts (as described below), with one expert assuming $x \in \treepath{h}$, and the other expert assuming $x \notin \treepath{h}$. This ensures that there always exists an expert for which all assumptions are correct. 
        }
    \end{enumerate}
    In greater detail, the contents of the state tuple $e = (S,u,H)$ represents the knowledge and assumptions of the expert as follows:
    \begin{itemize}
        \item[$\circ$]{
            $u \in \binarytree{d}$ -- This single node encodes everything the expert knows and assumes about which of the nodes labeled so far are on-path. Observe that if $v_1,v_2,\dots,v_k \in \binarytree{d}$ are nodes that are assumed to be on-path (and all these assumptions are consistent), then these $k$ assumptions can be represented succinctly by assigning $u = v_{i^*}$ where $v_{i^*}$ is the deepest node among $v_1,v_2,\dots,v_k$. Therefore, $u$ simply holds the deepest node in the tree that is known or assumed to be on-path. At the start of the algorithm, this value is initialized to be $u = \lambda$, because the root is known to be on-path regardless of the target function.
        }
        \item[$\circ$]{
            $S \subseteq \binarytree{d}$ -- the `danger zone', as described in \cref{section:danger-zone-minimization}. This is a collection that contains all nodes in the prefix $x_{\leq \tmax} = \paren*{x_1,x_2,\dots,x_{\tmax}}$ of the sequence to be classified that have not been labeled yet and \emph{might} be on-path for the true labeling function $h$ given what the expert knows and assumes so far. However, $S$ is not required to contain ancestors of nodes that are assumed to be on-path. Initially, $S$ equals the prefix $x_{\leq \tmax}$. As information accumulates, nodes that cannot be on-path are removed from $S$. For instance, if $x_i \in \binarytree{d}$ is assigned label $y_i \in \zo$ by the adversary, then any $(1-y_i)$-descendant of $x_i$ (including $x_i$ itself) may safely be removed from $S$.
        }
        \item[$\circ$]{
            $H \subseteq \zo^{\binarytree{d}}$ -- the version space of the experts, i.e., the collection of all functions that could be the correct labeling function given everything that the expert knows and assumes. Initially, $H$ contains all functions in $\cH$. As information accumulates, some functions are ruled out. Specifically, a function $h$ can be removed from $H$ for two reasons: (i) the adversary assigns a label $y\neq h(x)$ to some node $x \in \binarytree{d}$; (ii) the expert makes an assumption that some $x \in \binarytree{d}$ is on-path for the correct labeling function but $x \notin \treepath{h}$, or vice versa, the expert assumes that $x$ is off-path for the correct labeling function but $x \in \treepath{h}$.
        }
    \end{itemize}

\paragraph{Updates and splits.}
    An expert can be modified using the procedure $\ExpertExtendedUpdate(e, \cdot, \cdot)$. This procedure either returns a single modified tuple $(S,u,H)$ (in the first two return statements in the procedure), in which case we think of the expert as being \emph{updated}; or alternatively, the procedure returns two tuples $e_{\in} = (\Son,\uon, \Hon)$ and $e_{\notin} = (\Soff, \uoff, \Hoff)$ (in the \codehyperlink{line:return-update-split}{third return statement}), in which case we think of the expert as being \emph{split} into two experts. The expert $e_{\in}$ corresponds to adding an assumption that the most recently presented node $x_t$ is on-path for the correct labeling function, and $e_{\notin}$ corresponds to adding the opposite assumption.

\paragraph{Ancestry.}
    At the end of each iteration of the outer `for' loop in \cref{algorithm:upper-bound-main}, for each expert $e \in E_{t+1}$ there exists a unique \emph{ancestry} sequence $\ancestry{e} = (e_1,e_2,\dots,e_{t+1})$ such that $e_1 = \left(\{x_1,\dots,x_{\tmax}\}, \lambda, \cH\right)$ is the initial single expert that was created before the start of the outer `for' loop, $e_{t+1}=e$ is the latest version of the expert, and for each $i \in [t]$, the expert $e_{i+1}$ was created by an execution of $\ExpertBasicUpdate(e_i, \cdot, \cdot)$ possibly followed by an execution of $\ExpertExtendedUpdate$.\footnote{
        Note that in this paper, we use genealogical metaphors in two distinct contexts that should not be confused. First, as is customary, we use ``child'', ``parent'', ``ancestor'' and ``descendant'' to describe relations between nodes in the binary tree $\binarytree{d}$, which constitutes the domain of our hypothesis class. Separately from that, we use ``ancestor'' and ``descendant'' to describe relations between experts. 
        
        This overlap in terminology can partially be excused by the fact that the history of experts also forms a binary tree. Indeed, initially there is a single expert (the root of the tree), and experts can split into two, corresponding to a node having two children as in a binary tree. Seeing as experts cannot merge, the expert history corresponds precisely to a binary tree. (However, the domain $\binarytree{d}$ is a \emph{perfect} binary tree, whereas the binary tree corresponding to expert genealogy need not be balanced).

        To reduce confusion, we use $\treepath{\cdot}$ only for nodes in $\binarytree{d}$, and $\ancestry{\cdot}$ only for experts, even though these operators are mathematically equivalent (however, $\treepath{\cdot}$ is defined not only for nodes in $\binarytree{d}$ but also for functions $\binarytree{d} \to \zo$).
    }

\ifneurips
\begin{algorithmFloat}[ht]
\else
\subsubsection{Pseudocode}
\begin{algorithmFloat}[H]
\fi
    \begin{ShadedBox}
        \textbf{Assumptions:}
        \begin{itemize}
            \item{
                $d,n \in \bbN$, $\lambda$ is the empty string.
            }
            \item{
                $\cH \subseteq \zo^\binarytree{d}$ is the class that exists by \cref{lemma:hypothesis-class}.
            }
            \item{
                $x_1,x_2,\dots,x_n \in \binarytree{d}$ are points to be classified.
            }
        \end{itemize}

        \vspp
        \TransductiveLearner($\cH$, $d$, $(x_1,x_2,\dots,x_n)$):
        \vsp
        \begin{algorithmic}

            \State $t \gets 0$, $\tmax \gets \tmaxvalue$

            \State $e \gets \left(\{x_1,\dots,x_{\tmax}\}, \lambda, \cH\right)$
            \FixedWidthComment{The initial expert. An expert is defined by a $3$-tuple.}{21em}

            \State $w(e) \gets 1$
            \FixedWidthComment{Assign the initial expert a weight of $1$.}{21em}

            \State $E_1 \gets \{e\}$
            \FixedWidthComment{$E_t$ is the set of experts used for predicting $\hat{y}_t$.}{21em}

            \State $E_2,\dots,E_n,E_{n+1} \gets \varnothing$

            \vsp

            \For $t \gets 1,2,\dots,n$: 

                \vsp

                \State $\displaystyle
                \hat{y}_t \gets \indicator{%
                    \sum_{e \in E_t}w(S)\cdot\ExpertPrediction(e, x_t) 
                    \geq
                    \frac{1}{2}%
                }$
                \FixedWidthComment{A weighted majority, using \cref{algorithm:expert-predict}.}{11em}

                \vspp

                \State \textbf{send} prediction $\hat{y}_t$ to adversary
                
                \vsp

                \State \textbf{receive} correct label $y_t \in \zo$ from adversary
                \vspp

                \For $e \in E_t$:
                \FixedWidthComment{Update the experts.}{15em}
                    
                    \vsp

                    \State $e \gets \ExpertBasicUpdate(e, x_t, y_t)$
                    \FixedWidthComment{Remove functions that disagree with the label $y_t$ from the version space.}{15em}

                    \vsp
                    
                    \If $\ExpertPrediction(e, x_t) = y_t$:
                        \State $E_{t+1} \gets E_{t+1} \cup \{e\}$
                        \FixedWidthComment{If expert $e$ made a correct prediction, no further update is needed.}{15em}
                    
                    \Else:

                        \State $U \gets \ExpertExtendedUpdate(e, x_t, y_t)$
                        \FixedWidthComment{If $e$ made a mistake, update $e$ using \cref{algorithm:expert-extended-update}. This might cause $e$ to be split into two experts.}{10em}

                        \For $e' \in U$:
                            \State $E_{t+1} \gets E_{t+1} \cup \{e'\}$
                            \FixedWidthComment{Add updated expert(s) to $E_{t+1}$.}{15.1em}
                            \State $w(e') \gets w(e)/(2\cdot|U|)$
                            \FixedWidthComment{When $e$ makes a mistake, its weight is decreased by a factor of $2$ and then split equally between its descendants.}{15.1em}
                        \EndFor
                    \EndIf
                \EndFor

                \vsp

            \EndFor

        \end{algorithmic}
    \end{ShadedBox}
    \vspace{-1em}
    \caption{A transductive online learning algorithm that makes at most $\Osqrtd$ mistakes. It is a variant of the multiplicative weights algorithm that employs splitting experts. Namely, we start with a single expert, and when an expert makes a mistake it may split into two experts. The behavior of the experts is defined in \cref{algorithm:expert-predict,algorithm:expert-extended-update}.}
    \label{algorithm:upper-bound-main}
\end{algorithmFloat}

\begin{complexAlgorithmFloat}[ht]
\begin{subalgfloat}[ht]
    \begin{ShadedBox}
        \textbf{Assumptions:}
        \begin{itemize}
            \item{
                $d \in \bbN$, $x \in \binarytree{d}$.
            }
            \item{
                $e = (S,u, H)$ is a tuple that defines an expert:
                \begin{itemize}
                    \item[$\circ$]{
                        $S \subseteq \binarytree{d}$ -- a collection of nodes that could be on-path for the true labeling function given what the expert knows and assumes.
                    }
                    \item[$\circ$]{
                        $u \in \binarytree{d}$ -- the deepest node known or assumed to be on-path by the expert.
                    }
                    \item[$\circ$]{
                        $H \subseteq \zo^{\binarytree{d}}$ -- the collection of all functions that could be the correct labeling function given what the expert knows and assumes.
                    }
                \end{itemize}
            }
        \end{itemize}
        
        \vspp

        \ExpertPrediction($e$, $x$):
        \vsp
        \begin{algorithmic}
            \State $(S,u,H) \gets e$
            \FixedWidthComment{Unpack the state that defines the expert.}{17em}

            \vspp 

            \hypertarget{line:return-predict-halving}{}
            \If $|H| \leq 2^{\constA\sqrt{d}}$:
                \State \textbf{return} $\HalvingPredict(H, x)$
                \FixedWidthComment{Once $H$ becomes small enough, simulate the Halving algorithm (\cref{algorithm:halving}). [\cref{case:predict-halving}]}{17em}
            \EndIf

            \vspace*{-0.5em}

            \hypertarget{line:return-predict-on-path-descendant}{}
            \If $x \ancestor u$:
                \State \textbf{return} $b \in \zo$ such that $x \ancestor_b u$
                \FixedWidthComment{$u$ is assumed to be on-path. If $u$ is a $b$-decendant of $x$, then the correct label for $x$ must be $b$. [\cref{case:predict-on-path-descendant}]}{17em}
            \EndIf
            
            \vspp
            
            \hypertarget{line:return-predict-majority}{}
            \State \textbf{return} $\displaystyle 
            \indicator{
                |\{x' \in S: ~ x \ancestor_1 x' \}|
                >
                |S|/3
            }$
            \FixedWidthComment{Output some $b \in \zo$ such that more than $1/3$ of suspected on-path nodes are $b$-decendants of $x$, if such a $b$ exists. Otherwise (when at least $1/3$ of $S$ are non-descendants of $x$), output~$0$. [\cref{case:update-halving,case:update-S-decrease,case:update-xt-off-path,case:update-xt-on-path}]}{17em}
        \end{algorithmic}
    \end{ShadedBox}
    \vspace{-1em}
    \caption{A subroutine of \cref{algorithm:upper-bound-main} that defines how an expert makes predictions.}
    \label{algorithm:expert-predict}
\end{subalgfloat}

\begin{subalgfloat}[ht]
    \begin{ShadedBox}
        \textbf{Assumptions:} 
        \begin{itemize}
            \item{
                $x$, $e$, $S$, $u$, $H$ -- as in \cref{algorithm:expert-predict}.
            }
            \item{
                $y$ -- the correct label for $x$, as selected by the adversary.
            }
        \end{itemize}

        \vsppp
        
        \ExpertBasicUpdate($e$, $x$, $y$):
        \vsp
        \begin{algorithmic}
            \State $(S,u, H) \gets e$
            \FixedWidthComment{Unpack the state that defines the expert.}{17.5em}
            
            \vspp

            \hypertarget{line:update-H}{}
            \State $H \gets \HalvingUpdate(H, x, y)$
            \FixedWidthComment{Update the version space, as in the Halving algorithm (\cref{algorithm:halving}).}{17.5em}

            \State \textbf{return} $(S,u, H)$
        \end{algorithmic}
    \end{ShadedBox}
    \vspace{-1em}
    \caption{A subroutine of \cref{algorithm:upper-bound-main} that defines how an expert is updated each time that a label is selected by the adversary.}
    \label{algorithm:expert-basic-update}
\end{subalgfloat}

\begin{subalgfloat}[ht]
    \begin{ShadedBox}
        \textbf{Assumptions:} 
        \begin{itemize}
            \item{
                $d$, $x$, $e$, $S$, $u$, $H$ -- as in \cref{algorithm:expert-predict}.
            }
            \item{
                $y$ -- the correct label for $x$, as selected by the adversary.
            }
        \end{itemize}
        
        \vsppp
        
        \ExpertExtendedUpdate($e$, $x$, $y$):
        \vsp
        \begin{algorithmic}

            \State $(S,u, H) \gets e$
            \FixedWidthComment{Unpack the state that defines the expert.}{17.5em}
            
            \vspp

            \hypertarget{line:return-update-halving}{}
            \If $|H| \leq 2^{\constA\sqrt{d}}$:
                \FixedWidthComment{If the version space is small, we just simulate the Halving algorithm, so the update is complete. [\cref{case:update-halving}]}{17.2em}
                \vspace*{-2em}
                \State \textbf{return} $\{(S, u, H)\}$
            \EndIf

            \vspp
            
            \For $b \in \zo$:
                \State $~ S_{b} \gets \{x' \in S: ~ x \ancestor_{b} x' \}$
                \FixedWidthComment{Set of suspected on-path nodes that are $b$-descendant of $x$.}{17.5em}
            \EndFor

            \vsp

            \hypertarget{line:return-update-S-decrease}{}
            \If $|S_{(1-y)}| > |S|/3$:
                \vsp
                \State $S' \gets S \setminus S_{(1-y)}$
                \FixedWidthComment{At least $1/3$ of suspected on-path nodes were $b$-decendants of $x$, and therefore the expert predicted label $\hat{y} = b$. But the correct label was $y = 1-b$. Remove all $b$-descendants of $x$ from~$S$. [\cref{case:update-S-decrease}]}{21em}
                \vspace*{-3em}
                \State \textbf{return} $\{(S',u,H)\}$
                \vspp
                \vspp
                \vspp
            \Else :
                \hypertarget{line:return-update-split}{}
                \State $\Soff \gets S$; ~ $\uoff \gets u$
                \FixedWidthComment{Split $e$ in two. First, construct $\eoff$ to be an updated version of $e$ after adding the assumption that $x \notin \treepath{h}$ for the correct labeling function $h$.}{17.5em}
                \vspace*{-3em}
                \State $\Hoff = \left\{h \in H: ~ x \notin \treepath{h}\right\}$
                \State $\eoff \gets (\Soff,\uoff,\Hoff)$
            
                \vspp

                \vsp

                \State $\Son \gets S_0 \cup S_1$
                \FixedWidthComment{Next, construct $\eon$ to be an updated version of $e$ adding the assumption $x \in \treepath{h}$. $\Son$~contains only nodes that are descendants of~$x$.}{17.5em}
                
                \vsp

                \State $\uon \gets u$
                \FixedWidthComment{$\uon$ represents updating the prior assumption that $u$ is on path by adding that $x$ is also on path.}{17.3em}
                \vspace*{-2em}
                \If $\uon \ancestor x$:
                    \State $\uon \gets x$
                \EndIf

                \vsp

                \State $\Hon = \left\{h \in H: ~ x \in \treepath{h}\right\}$
                \FixedWidthComment{$\Hon$ is obtained by updating the version space to include only function where $x$ is on path.}{17.3em}
                \vspace*{-0.5em}
                \State $\eon \gets (\Son,\uon,\Hon)$

                \vsp

                \State \textbf{return} $\{\eoff,\eon\}$
                \FixedWidthComment{[\cref{case:update-xt-off-path,case:update-xt-on-path}]}{17.5em}
            \EndIf
        \end{algorithmic}
    \end{ShadedBox}
    \vspace{-1em}
    \caption{A subroutine of \cref{algorithm:upper-bound-main} that defines how an expert is updated (and possibly split into two) when it makes a mistake.}
    \label{algorithm:expert-extended-update}
\end{subalgfloat}
\end{complexAlgorithmFloat}

\begin{algorithmFloat}[ht]
    \begin{ShadedBox}
        \textbf{Assumptions:}
        \begin{itemize}
            \item{$\cX$ a set, $k \in \bbN$.}
            \item{$\cH \subseteq \zo^\cX$ is a finite hypothesis class.}
            \item{$x,x_1,\dots,x_k \in \cX$, $y \in \zo$.}
        \end{itemize}

        \vsppp

        \Halving($\cH$, $(x_1,x_2,\dots,x_k)$):
        \vsp
        \begin{algorithmic}

            \State $\cH_1 \gets \cH$

            \vsp

            \For $i \in [k]$:
                
                \vsp

                \State $\hat{y}_i \gets \HalvingPredict(\cH, x_i)$

                \vsp

                \State \textbf{send} prediction $\hat{y}_i$ to adversary
                
                \vsp

                \State \textbf{receive} correct label $y_i \in \zo$ from adversary

                \vsp

                \State $\cH_{i+1} \gets \HalvingUpdate(\cH_i, x_i, y_i)$
            \EndFor
        \end{algorithmic}

        \vsppp

        \HalvingPredict($\cH$, $x$):
        \vsp
        \begin{algorithmic}

            \State \textbf{return} $\indicator{%
                \frac{1}{|\cH|}\sum_{h \in \cH}h(x) 
                \geq
                \frac{1}{2}%
            }$
            
        \end{algorithmic}

        \vsppp

        \HalvingUpdate($\cH$, $x$, $y$):
        \vsp
        \begin{algorithmic}
            \State \textbf{return} $\big\{h \in \cH: ~ h(x) = y \big\}$
        \end{algorithmic}
    \end{ShadedBox}
    \vspace{-1em}
    \caption{This is the well-known halving algorithm. The experts in \cref{algorithm:expert-predict,algorithm:expert-extended-update} simulate this algorithm once their version space becomes small enough.}
    \label{algorithm:halving}
\end{algorithmFloat}

\subsection{Analysis}

In this section we prove our main result, \cref{theorem:separation}.

\subsubsection{Assumption-Consistent Expert}

Occasionally, when an expert is updated, it makes an assumption about whether the most-recently presented node $x_t$ is on-path or off-path with respect to the true labeling function $h$. In these updates, the expert is split into two: one expert assumes that $x_t \in \treepath{h}$, and the other assumes $x_t \notin \treepath{h}$. Clearly, by splitting into two in this manner, we preserve the invariant that the set of experts always contains a `vindicated' expert $e^*$ such that all the assumptions made by $e^*$ are correct. This simple observation is made formal in the following definition and claim. 

\begin{definition}[Assumption consistency]
    \label{definition:assumption-consistent}
    For an expert $e \in E_{t+1}$ with $\ancestry{e} = (e_1,e_2,\dots,e_{t+1})$, and an index $i \in [t]$, we say that the $i \to (i+1)$ update of $e$ was \ul{assumption-consistent} with a function $h: ~ \binarytree{d} \to \zo$ if one of the following conditions holds:
    \begin{itemize}
        \item{
            $e_{i+1} = \ExpertBasicUpdate(e_i, x_i, y_i)$; or
         }
        \item{
            $e_{i+1}$ was the single expert returned when executing $\ExpertExtendedUpdate(e_i', x_i, y_i)$ for $e_i' = \ExpertBasicUpdate(e_i, x_i, y_i)$; or
        }
        \item{
            Executing $\ExpertExtendedUpdate(e_i', x_i, y_i)$ with $e_i' = \ExpertBasicUpdate(e_i, x_i, y_i)$ returned two experts $(\Son,\uon,\Hon)$ and $(\Soff, \uoff,\Hoff)$ (as in the \codehyperlink{line:return-update-split}{third return statement}), and furthermore,
            \begin{equation}
                \label{eq:assumption-consistency}
                e_{i+1} =
                \left\{
                \begin{array}{ll}
                    (\Son,\uon,\Hon) & x_i \in \treepath{h} \\
                    (\Soff, \uoff,\Hoff)   & x_i \notin \treepath{h}.
                \end{array}
                \right.
            \end{equation}
        }
    \end{itemize}
    We say that an expert $e \in E_{t+1}$ is \ul{assumption-consistent} with $h$ if for all $i \in [t]$, the $i \to (i+1)$ update of $e$ was assumption-consistent with $h$.
\end{definition}

\begin{claim}[Existence of assumption-consistent expert]
    \label{claim:assumption-consistent-expert-exists}
    Let $d,n,t \in \bbN$,  $t \leq n$, let $\cH \subseteq \zo^\binarytree{d}$, let $x_1,\dots,x_n \in \binarytree{d}$, and let $h: ~ \binarytree{d} \to \zo$. Consider an execution of 
    \[
        \TransductiveLearner\paren*{\cH, d, \paren*{x_1,x_2,\dots,x_n}}
    \]
    as in \cref{algorithm:upper-bound-main}. Then, at the end of the $t$-th iteration of the outer `for' loop in \TransductiveLearner, there exists a unique expert $e^*_{t+1} \in E_{t+1}$ that is assumption-consistent with $h$.
\end{claim}

\begin{proof}
    We prove by induction that, for all $s \in [t+1]$, $E_{s}$ contains a unique expert that is assumption-consistent with $h$. The base case $s = 1$ is clear, because $E_1$ contains only a single expert that was never modified. For the induction step, let $e_s^*$ be the unique assumption-consistent expert in $E_s$, and consider the $s \to (s+1)$ update. Notice that by \cref{definition:assumption-consistent},
    \begin{itemize}
        \item{
            For all $e \in E_s \setminus \{e^*_s\}$, every expert $e' \in E_{s+1}$ such that $e'$ was created from $e$ by executing $\ExpertBasicUpdate(e_s, x_s, y_s)$ possibly followed by an execution of $\ExpertExtendedUpdate$ is not assumption-consistent with $h$; and
        }
        \item{
            Either $\ExpertBasicUpdate(e^*_s, x_s, y_s) \in E_{s+1}$ and $\ExpertExtendedUpdate(e_s^*, x_s, y_s)$ is not executed ($e^*_s$ is added to $E_{s+1}$ with just a basic update), or precisely one of the experts that were created from $e^*_s$ by executing $\ExpertExtendedUpdate$ and added to $E_{s+1}$ is assumption-consistent with $h$.
        }
    \end{itemize}
    Seeing as the $s \to (s+1)$ update executes $\ExpertBasicUpdate$ and $\ExpertExtendedUpdate$ at most once for each $e \in E_s$, it follows that $E_{s+1}$ contains precisely one expert that is assumption-consistent with $h$.
\end{proof}

An expert $e = (S,u,H)$ that is assumption-consistent with the correct labeling function enjoys two simple properties. The first property is that the node $u$ in the expert encodes correct information about which previously seen nodes are on-path for the correct labeling function.

The second property is that the set $S$ contains all future nodes that are on-path for the correct labeling function and are also deeper in the tree than all nodes assumed to be on-path so far. These two properties are formalized in the following claim.

\begin{claim}[Properties of assumption-consistent expert]
    \label{claim:assumption-consistent-expert-properties}
    Let $d,n,t \in \bbN$,  $t \leq n+1$, let $\cH \subseteq \zo^\binarytree{d}$, let $x_1,\dots,x_n \in \binarytree{d}$. Consider an execution of 
    \[
        \TransductiveLearner\paren*{\cH, d, \paren*{x_1,x_2,\dots,x_n}}
    \]
    as in \cref{algorithm:upper-bound-main}. Assume that the adversary selects labels $y_1,y_2,\dots,y_n \in \zo$ that are consistent with some function $h: ~ \binarytree{d} \to \zo$. Let $e_t^* = (S_t^*,u_t^*,H_t^*) \in E_t$ be the unique expert in $E_t$ that is assumption-consistent with $h$.\footnote{Recall that $e_t^*$ exists by \cref{claim:assumption-consistent-expert-exists}.} Then the following two properties hold:
    \begin{enumerate}
        \item{
            \label{item:u-on-path}
            $u_t^* \in \treepath{h}$.
        }
        \item{
            \label{item:S-contains-on-path}
            $\set*{
                x \in \set*{x_t,x_{t+1},\dots,x_{\tmax}}:
                ~ 
                x \in \treepath{h} 
                ~ 
                \land
                ~ 
                x \not\ancestor u_t^*
            }
            \subseteq
            S_t^*$.
        }
    \end{enumerate}
\end{claim}

\begin{proof}[Proof of \cref{claim:assumption-consistent-expert-properties}]
    The proof proceeds by induction on $t$. For the base case $t=1$, $E_1$ contains a single expert $e_1^* = (S_1^*,u_1^*,H_1^*)$ where $u_1^* = \lambda$ is the root of $\binarytree{d}$. Indeed, $\lambda \in \treepath{h}$ for any function $h: ~ \binarytree{d} \to \zo$. This establishes the base case for \cref{item:u-on-path}. Additionally, $S_1^* = \set*{x_1,x_2,\dots,x_{\tmax}}$, satisfying the base case for \cref{item:S-contains-on-path}.

    For the induction step, we assume that the claim holds for some integer $t = i$, and show that it holds for $t = i+1$ as well. First, we establish \cref{item:u-on-path}. If $e_{i+1}^* = \ExpertBasicUpdate(e_i^*, x_i, y_i)$, then the claim is immediate because $u_{i+1}^* = u_i^* \in \treepath{h}$. Otherwise, by \cref{definition:assumption-consistent} and the first \codehyperlink{line:return-update-halving}{first} \codehyperlink{line:return-update-S-decrease}{two} return statements in $\ExpertExtendedUpdate$, either $e_{i+1}^* = (S_{i+1}^*, u_{i+1}^*, H_{i+1}^*)$ has $u_{i+1}^* = u_i^* \in \treepath{h}$, in which case the claim is immediate, or else $e_{i+1}^*$ satisfies \cref{eq:assumption-consistency}, namely,
    \[
        e_{i+1}^* =
        \left\{
        \begin{array}{ll}
            (\Son,\uon,\Hon) & x_i \in \treepath{h} \\
            (\Soff, \uoff,\Hoff)   & x_i \notin \treepath{h}.
        \end{array}
        \right.
    \]
    As defined in $\ExpertExtendedUpdate$, $\uon$ is equal either to $u_i^*$ or to $x_i$, so if $x_i \in \treepath{h}$ then 
    \[
    u_{i+1}^* = \uon \in \{u_i^*,x_i\}\subseteq \treepath{h}.
    \]
    On the other hand, if $x_i \notin \treepath{h}$ then we get $u_{i+1}^* = \uoff = u_i^* \in \treepath{h}$. We see that in all cases, $u_{i+1}^* \in \treepath{h}$ as desired. This concludes the proof of \cref{item:u-on-path}.

    For \cref{item:S-contains-on-path}, again, if $e_{i+1}^* = \ExpertBasicUpdate(e_i^*, x_i, y_i)$, then the claim is immediate because $S_{i+1}^* = S_i^*$ and $u_{i+1}^* = u_i^*$. Otherwise, consider the various ways in which $u_{i+1}^*$ and $S_{i+1}^*$ can be assigned by $\ExpertExtendedUpdate$. In the \codehyperlink{line:return-update-halving}{first return statement}, $u_{i+1}^* = u_i^*$ and $S_{i+1}^* = S_i^*$, and the claim is immediate.
    
    The \codehyperlink{line:return-update-S-decrease}{second return statement} assigns $u_{i+1}^* = u_i^*$ and $S_{i+1}^* = S_i^* \setminus S_{1-y_i}$, where $S_{1-y_i}$ is the set of $(1-y_i)$-descendants of $x_i$ (including $x_i$ itself). Notice that regardless of whether $x_i$ is on-path for the correct labeling function $h$ or not, none of the $(1-y_i)$-descendants of $x_i$ (except possibly $x_i$ itself) can be on-path for $h$, because $h$ assigns a label $y_i$ to $x_i$. And seeing as \cref{item:S-contains-on-path} only requires that $S_{i+1}^*$ contain nodes from $\set*{x_{i+1},x_{i+2},\dots,x_{\tmax}}$, it is also safe to remove $x_i$. Therefore, removing $S_{1-y_i}$ preserves \cref{item:S-contains-on-path}.

    For the \codehyperlink{line:return-update-split}{third return statement}, there are two possibilities. The first possibility is that $u_{i+1}^* = \uoff = u_i^*$ and $S_{i+1}^* = \Soff = S_i^*$, in which case the claim is immediate. The second possibility assigns $u_{i+1}^* = \uon$, and $S_{i+1}^* = \Son = S_0 \cup S_1$, namely, $S_{i+1}^*$ is constructed by removing the non-descendants of $x_i$ from~$S_i^*$. By \cref{eq:assumption-consistency}, this happens when $x_i \in \treepath{h}$, so all non-descendants of $x_i$ or either off-path for $h$, or they are ancestors of $x_i$. Seeing as $x_i \in \treepath{h}$ and $u_i^* \in \treepath{h}$, and $\uon$ is the deeper node between these two, any node that is an ancestor of $x_i$ is also an ancestor of $u_{i+1}^* = \uon$. Thus, all the nodes removed or either  off-path for $h$, or they are ancestors of $u_{i+1}^*$, satisfying \cref{item:S-contains-on-path}. (Similarly, any node that is an ancestor of $u_i^*$ is also an ancestor of $u_{i+1}^*$, so we do not need to add any new nodes to $S_{i+1}^*$ that are not included in $S_i^*$.)
    
    We see that in all cases, \cref{item:S-contains-on-path} is preserved, as desired.
\end{proof}

\subsubsection{Transition to Halving}

\begin{claim}
    \label{claim:transition-to-halving}
    Let $d,n,t \in \bbN$, $d \geq \constHalvingTransitionMinimum$, let $\cH \subseteq \zo^\binarytree{d}$, and let $x_1,\dots,x_n \in \binarytree{d}$. Consider an execution of 
    \[
        \TransductiveLearner(\cH, (x_1,x_2,\dots,x_n))
    \]
    as in \cref{algorithm:upper-bound-main}. Let $t > \tmax =  \tmaxvalue$ and let $e = (S, u, H) \in E_t$ be an expert. Then 
    \[
        |H| \leq 2^{\constA\sqrt{d}}.
    \]
\end{claim}

\begin{proof}[Proof of \cref{claim:transition-to-halving}]
    Assume for contradiction that $\left|H\right| > 2^{\constA\sqrt{d}}$. 
    Let $H' \subseteq H$ be an arbitrary subset of size $2^{\constA\sqrt{d}}+1$. Let 
    \[
        P = \cup_{h \in H'}\treepath{h}.
    \]
    Seeing as each root-to-leaf path contains $d+1$ nodes,
    \begin{equation}
        \label{eq:P-upper-bound}
        |P| \leq |H'| \cdot (d+1) \leq \left(2^{\constA\sqrt{d}}+1\right)\cdot(d+1) \leq d2^{\constA\sqrt{d} + 1}.
    \end{equation}

    Let $y_1,y_2,\dots,y_t$ be the labels provided by the adversary in the first $\tmax$ iterations. The \codehyperlink{line:update-H}{line} in $\ExpertBasicUpdate$ constructing $H$ using $\HalvingUpdate(H, x, y)$ ensures that
    \begin{equation}
        \label{eq:H-is-consistent}
        \forall h \in H 
        ~
        \forall i \in [\tmax]:
        ~
        h(x_i) = y_i.
    \end{equation}
    Consider two cases:
    \begin{caselist}
        \item{
            \label{case:majority_label_zero}
            $\sum_{i = 1}^{\tmax} y_i \leq \tmax/2$. Then the set
            \[
                X_0 = \{x_i: ~ i \in [\tmax] ~ \land ~ y_i = 0\}
            \]
            has cardinality $|X_0| \geq \tmax/2$. Let $X_0' = X_0 \setminus P$. By \cref{eq:P-upper-bound},
            \begin{equation}
                \label{eq:x0-prime-lower-bound}
                \left| X_0' \right| 
                \geq 
                \frac{\tmax}{2} - d  2^{\constA\sqrt{d} + 1}
                =
                2^{\constC\sqrt{d}} - d  2^{\constA\sqrt{d} + 1}.
            \end{equation}
            From the choice of $X_0'$, the inclusion $H' \subseteq H$, and \cref{eq:H-is-consistent}, 
            \begin{equation}
                \label{eq:class-property-holds-for-0}
                \forall h \in H' ~ \forall x \in X_0': ~ x \notin \treepath{h} ~ \land ~ h(x) = 0.
            \end{equation}
            Seeing as $|H'| > 2^{\constA \sqrt{d}}$, \cref{eq:class-property-holds-for-0} and \cref{item:class-property-for-0} from \cref{lemma:hypothesis-class} imply that 
            \begin{equation}
                \label{eq:x0-prime-upper-bound}
                |X_0'| \leq 2^{\constA \sqrt{d}}.
            \end{equation}
            Combining \cref{eq:x0-prime-upper-bound,eq:x0-prime-lower-bound} yields
            \begin{align*}
                2^{\constA \sqrt{d}}
                &\geq 
                |X_0'|
                \geq 
                2^{\constC\sqrt{d}} - d  2^{\constA\sqrt{d} + 1}
                \\
                &\geq
                2^{\constC\sqrt{d}-1}
                \tagexplain{$d \geq \constHalvingTransitionMinimum$},
            \end{align*}
            which is a contradiction.
        }
        \item{
            \label{case:majority_label_one}
            $\sum_{i = 1}^{\tmax} y_i > \tmax/2$. A similar argument gives a contradiction by defining
            \[
                X_1 = \{x_i: ~ i \in [\tmax] ~ \land ~ y_i = 1\}, ~~ \text{and} ~~ X_1' = X_1\setminus P.
            \]
            As before,
            \begin{equation}
                \label{eq:x1-prime-lower-bound}
                \left| X_1' \right| 
                \geq 
                \frac{\tmax}{2} - d  2^{\constA\sqrt{d} + 1}
                \geq 
                2^{\constC\sqrt{d}} - d  2^{\constA\sqrt{d} + 1}.
            \end{equation}
            for all $d \in \bbN$. However, $|H'| > 2^{\constA \sqrt{d}}$ and \cref{item:class-property-for-1} imply that
            \begin{equation}
                \label{eq:x1-prime-upper-bound}
                |X_1'| < \constB\sqrt{d},
            \end{equation}
            which is a contradiction.
            \qedhere
        }
    \end{caselist}
\end{proof}

\subsubsection{Performance of Best Expert}

\begin{claim}[Existence of expert with large weight]
    \label{claim:good-expert-exists}
    Let $d,n \in \bbN$, $d \geq \constHalvingTransitionMinimum$, let $\cH \subseteq \zo^\binarytree{d}$, and let $x_1,\dots,x_n \in \binarytree{d}$. Consider an execution of 
    \[
        \TransductiveLearner(\cH, (x_1,x_2,\dots,x_n))
    \]
    as in \cref{algorithm:upper-bound-main}.
    Then, at the end of the execution, there exists $e \in E_{n+1}$ such that
    \begin{equation}
        \label{eq:expert-with-large-weight}
        w(e) \geq 2^{-\constUpperBound \sqrt{d}}.
    \end{equation}
\end{claim}

Note that the lower bound in \cref{eq:expert-with-large-weight} does not depend on $n$. 

\begin{proof}
    Fix a hypothesis $h \in \cH$ such that $h(x_t) = y_t$ for all $t \in [n]$ (such an $h$ exists because the adversary must always select a realizable label).

    By \cref{claim:assumption-consistent-expert-exists}, there exists $e_{n+1}^* \in E_{n+1}$ that is assumption-consistent with $h$. Let $\ancestry{e_{n+1}^*} = (e_1^*,e_2^*,\dots,e_{n+1}^*)$. We argue that this ancestry sequence makes few mistakes. Specifically, for each $t \in [n]$, let $\hat{y}^*_t = \ExpertPrediction(e_t^*, x_t)$. We claim that
    \[
        m := \sum_{t = 1}^n \indicator{\hat{y}^*_t \neq y_t} \leq  \constD \sqrt{d}.
    \]
    Indeed, let $B = \{t \in [n]: ~ \hat{y}^*_t \neq y_t\}$ be the set of $m$ indices where a mistake was made. 
    For each $t \in B$, let $e_t^* = (S,u,H)$, and note that each $t \in B$ has a corresponding execution of $\ExpertPrediction(e_t^*, x_t)$, and an execution of $e_t' = \ExpertBasicUpdate(e_t^*, x_t, y_t)$ followed by $\ExpertExtendedUpdate(e_t', x_t, y_t)$ that produces $e_{t+1}^*$ ($\ExpertExtendedUpdate$ is executed because $t \in B$, i.e., a mistake was made).
    We partition the indices in $B$ into six cases (six disjoint sets), and bound the number of indices that fall in each.
    \begin{caselist}
        \item{
            \label{case:predict-halving}
            \emph{The execution of $\ExpertPrediction(e_t^*, x_t)$ exited via the \codehyperlink{line:return-predict-halving}{first return statement} in that procedure.} This happens once $|H|\leq 2^{\constA\sqrt{d}}$, and from that point on, the expert and all subsequent experts in the ancestry are exactly simulating the $\Halving$ algorithm (\cref{algorithm:halving}) in both  predictions and updates. Hence, by \cref{claim:halving}, $B$ contains at most $\mI = \constA\sqrt{d}$ such indices.
        }
        \item{
            \label{case:predict-on-path-descendant}
            \emph{The execution of $\ExpertPrediction(e_t^*, x_t)$ exited via the \codehyperlink{line:return-predict-on-path-descendant}{second return statement} in that procedure.} In particular $x \ancestor u$, and the predicted label was $\hat{y}^*_t = b \in \zo$ such that $x_t \ancestor_b u$. Because $e_t^*$ is assumption-consistent with $h$, \cref{item:u-on-path} in \cref{claim:assumption-consistent-expert-properties} implies that $u \in \treepath{h}$. Namely, we see that $u$ is a $b$-descendant of $x_t$ and $u \in \treepath{h}$. It follows that $\hat{y}^*_t = b = h(x_t) = y_t$. So no mistakes are made in \cref{case:predict-on-path-descendant}, and the number of indices $t \in B$ that belong to \cref{case:predict-on-path-descendant} is simply $\mII = 0$.
        }
    \end{caselist}
    In the remaining cases, we assume that $\ExpertPrediction(e_t^*, x_t)$ exited via the \codehyperlink{line:return-predict-majority}{third return statement} in that procedure, so the prediction was 
    \begin{equation}
        \label{eq:y-hat-eq-indicator}
        \hat{y}^*_t = \indicator{
            |S_1|
            >
            |S|/3
        },
    \end{equation}
    where $S_1 = \{x' \in S: ~ x_t \ancestor_1 x' \}$. These cases are as follows.
    \begin{caselist}
        \setcounter{caselisti}{2}
        \item{
            \label{case:update-halving}
            \emph{The execution of $\ExpertExtendedUpdate(e_t', x_t, y_t)$ exited via the \codehyperlink{line:return-update-halving}{first return statement} in that procedure.} Namely, after the update, the resulting expert $e_{t+1}^*$ has $|H| \leq 2^{\constA\sqrt{d}}$. However, because we are not in \cref{case:predict-halving}, at the beginning of the iteration expert $e_t^*$ had $|H| > 2^{\constA\sqrt{d}}$. Seeing as the cardinality of $H$ decreases monotonically throughout the ancestry $e_1^*,\dots,e_{n+1}^*$, this type of mistake can happen at most $\mIII = 1$ times.
        }
        \item{
            \label{case:update-S-decrease}
            \emph{The execution of $\ExpertExtendedUpdate(e_t', x_t, y_t)$ exited via the \codehyperlink{line:return-update-S-decrease}{second return statement} in that procedure.} In this case, $|S_{(1-y_t)}| > |S|/3$, and $e_{t+1}^* = (S',u,H)$ with $S' = S \setminus S_{1-y_t}$. So $|S'| < 2|S|/3$. Namely, the update causes the cardinality of the set $S$ to be multiplied by a factor of at most $2/3$ and it  strictly decreases. Seeing as the initial cardinality is $\tmax$, and cardinalities are integers, the number of times this can happen is at most 
            \begin{equation}
                \label{eq:m-iii-bound}
                \mIV = \frac{\log \tmax}{\logf{3/2}} + 1 = \frac{\constC\sqrt{d}}{\logf{3/2}} + 1.
            \end{equation}
        }
    \end{caselist}
    In the remaining cases, we assume that the execution of $\ExpertExtendedUpdate(e_t^*, x_t, y_t)$ exited via the \codehyperlink{line:return-update-split}{third return statement} in that procedure. This implies that 
    \begin{equation}
        \label{eq:minority-decision}
        |S_{\hat{y}^*_t}| \leq |S|/3
    \end{equation}
    Combining this with \cref{eq:y-hat-eq-indicator}, it follows $\hat{y}^*_t = 0$ and therefore $y_t = 1$. The remaining cases are as follows.
    \begin{caselist}
        \setcounter{caselisti}{4}
        \item{
            \label{case:update-xt-on-path}
            \emph{$x_t \in \treepath{h}$}. Let $e_t^* = (S, u, H)$. 
            Seeing as $|H| > 2^{\constA\sqrt{d}}$ (because we are not in \cref{case:predict-halving}), \cref{claim:transition-to-halving} (with the assumption $d \geq \constHalvingTransitionMinimum$) implies that $t \leq \tmax$. 
            By \cref{item:S-contains-on-path} of \cref{claim:assumption-consistent-expert-properties},
            the facts $x_t \not\ancestor u$ (we are not in \cref{case:predict-on-path-descendant}) 
            and $x_t \in \treepath{h}$ imply that $x_t \in S$. In particular, $S$ is not empty.
            
            Because the $t \to (t+1)$ update of $e_{t+1}^*$ was assumption-consistent with $h$, \cref{eq:assumption-consistency} implies that $e_{t+1}^* = (\Son,\uon,\Hon)$, with $\Son = S_0 \cup S_1$. Observe that 
            \begin{itemize}
                \item{
                    $|S_0| \leq |S|/3$ (plugging $\hat{y}^*_t = 0$ into \cref{eq:minority-decision}); and
                }
                \item{
                    $|S_1| \leq |S|/3$ (because otherwise, by \cref{eq:y-hat-eq-indicator}, the prediction would have been $\hat{y}^*_t = 1$).
                }
            \end{itemize}
            Therefore, 
            \begin{equation}
                \label{eq:s-prime-small}
                |\Son| \leq |S_0| + |S_1| \leq 2|S|/3.
            \end{equation}
            As in \cref{case:update-S-decrease}, combining \cref{eq:s-prime-small} and the fact that $S$ is not empty imply an upper bound $\mV$ on the number of times \cref{case:update-xt-on-path} can happen, with the bound being the same number $\mV = \mIV$ as in \cref{eq:m-iii-bound}.
        }
        \item{
            \label{case:update-xt-off-path}
            \emph{$x_t \notin \treepath{h}$}. So $(x_t,y_t)$ is a pair such that $x_t \notin \treepath{h}$ and $y_t=1$. Assume for contradiction that this type of mistake can happen strictly more than 
            \[
                \mVI = \constB \sqrt{d}
            \]
            times. Let $t_1,t_2,\dots,t_{\mVI}$ be the indices of the first $\mVI$ iterations of the outer `for' loop of $\TransductiveLearner$ in which this type of mistake happened. Note that if at the end of iteration $t_{\mVI}$, we had expert $e^*_{t_{\mVI}+1} = (S_{t_{\mVI}+1},u_{t_{\mVI}+1},H_{t_{\mVI}+1})$ such that $|H_{t_{\mVI}+1}| \leq 2^{\constA\sqrt{d}}$, then from that point onwards, the expert would be simulating the halving algorithm, and in particular, it would not make any further mistake of the type in \cref{case:update-xt-off-path} (all subsequent mistakes would belong to \cref{case:predict-halving}). Hence, by the assumption that strictly more than $\mVI$ mistakes were made, it follows that $|H_{t_{\mVI}+1}| > 2^{\constA\sqrt{d}}$. 
            Let 
            \[
                H^* = \big\{
                    h' \in \cH: ~ 
                        \left(
                            \forall t \in [\mVI]: ~ 
                                h'(x_{t_t}) = 1
                                ~ \land ~
                                x_t \notin \treepath{h'}
                        \right)
                \big\}.
            \]
            Because $e^*_{t_{\mVI}+1}$ is assumption-consistent with $h$, and from the construction of $H_{t_{\mVI}+1}$ using $\Hon$ and $\Hoff$ in $\ExpertExtendedUpdate$, it follows that $H_{t_{\mVI}+1} \subseteq H^*$. 
            So there exist collections $H^* \subseteq \cH$ and $X = \{x_{t_t}: ~ t \in [\mVI]\} \subseteq \binarytree{d}$ such that
            \begin{itemize}
                \item{
                    $|H^*| \geq |H_{t_{\mVI}+1}| > 2^{\constA\sqrt{d}}$,
                }
                \item{
                    $|X| = \mVI = \constB \sqrt{d}$,
                }
                \item{
                    $\forall h' \in H^* ~ \forall x \in X: ~ h'(x) = 1$.
                }
                \item{
                    $\forall h' \in H^* ~ \forall x \in X: ~ x \notin \treepath{h'}$.
                }
            \end{itemize}
            This is a contradiction to the choice of $\cH$, specifically, to \cref{item:class-property-for-1} in \cref{lemma:hypothesis-class}.
        }
    \end{caselist}
    Thus, combining the analyses of all cases, we see that the number of mistakes made by the $\ancestry{e_{n+1}^*}$ is at most 
    \begin{align*}
        m 
        &\leq 
        \mI + \mII + \mIII + \mIV + \mV + \mVI 
        \\
        &\leq 
        \constA\sqrt{d}
        +
        0
        +
        1
        +
        \paren*{\frac{\constC\sqrt{d}}{\logf{3/2}} + 1}
        +
        \paren*{\frac{\constC\sqrt{d}}{\logf{3/2}} + 1}
        +
        \constB \sqrt{d}
        \\
        &\leq 
        \constD \sqrt{d}.
    \end{align*}
    The weights satisfy
    \[
        w(e_{t+1}^*) 
        ~
        \left\{
        \begin{array}{ll}
           = ~ w(e_t^*)  & ~~ \hat{y}^*_t = y_t \\
           \geq ~ \frac{1}{4}\cdot w(e_t^*) & ~~ \hat{y}^*_t \neq y_t.
        \end{array}
        \right.
    \]
    This implies that $w(e_{n+1}^*) \geq w(e_1^*)\cdot\prod_{t = 1}^n 4^{-\indicator{\hat{y}_i \neq y_i}} = w(e_1^*) \cdot 4^{-m} \geq 4^{-\constD \sqrt{d}} = 2^{-\constUpperBound \sqrt{d}}$, as desired.
\end{proof}

\subsubsection{Multiplicative Weights Mistake Bound}

\begin{claim}[Mistake bound for multiplicative weights]
    \label{claim:multiplicative-weights-mistake-bound}
    Let $d,n \in \bbN$, let $\alpha > 0$, let $\cH \subseteq \zo^\binarytree{d}$, and let $x_1,\dots,x_n \in \binarytree{d}$. Consider an execution of 
    \[
        \TransductiveLearner(\cH, (x_1,x_2,\dots,x_n))
    \]
    as in \cref{algorithm:upper-bound-main}. 
    Assume that at the end of the execution, there exists $e^* \in E_{n+1}$ such that
    \[
        w(e^*) \geq 2^{-\alpha}.
    \]
    Then $\TransductiveLearner$ makes at most $\alpha$ mistakes.
\end{claim}

\begin{proof}[Proof of \cref{claim:multiplicative-weights-mistake-bound}]
    For all $i \in [n+1]$, let $w(E_i) = \sum_{e \in E_i}w(e)$. 
    For each $i \in [n]$, if $\hat{y}_i \neq y_i$, then $w(E_{i+1}) \leq w(E_i)/2$. Hence, if $\TransductiveLearner$ makes $m$ mistakes, then by induction
    \[
        w(E_{n+1}) \leq w(E_1)\cdot\prod_{t = 1}^n 2^{-\indicator{\hat{y}_i \neq y_i}} = 2^{-m}\cdot w(E_1).
    \]
    So
    \[
        2^{-\alpha} \leq w(e^*) \leq \sum_{e \in E_{n+1}}\!w(e) = w(E_{n+1}) \leq 2^{-m}\cdot w(E_1) = 2^{-m}.
    \]
    We conclude that
    \[
    m \leq \alpha,
    \]
    as desired.
\end{proof}

\subsection{Proof}

\begin{proof}[Proof of \cref{theorem:separation}]
    Fix an integer $d \geq \constUpperBoundMinimum$. Let $\cH \subseteq \zo^{\binarytree{d-1}}$ be the class constructed by invoking \cref{lemma:hypothesis-class} for the integer $d-1 \geq \constClassMinimum$. We argue that this class satisfies the requirements of \cref{theorem:separation}. 

    By construction, $\cH$ is a class of Littlestone dimension precisely $d$. By \cref{theorem:ld-is-upper-bound}, this implies the equality in \cref{item:main-theorem-lower-bound}.
    
    We now show the upper bound in \cref{item:main-theorem-upper-bound}. We argue that $\TransductiveLearner$ (\cref{algorithm:upper-bound-main}) satisfies this upper bound. 
    By \cref{claim:good-expert-exists}, at the end of the execution of $\TransductiveLearner$ there exists an expert $e \in E_{n+1}$ such that $w(e) \geq 2^{-\constUpperBound \sqrt{d}}$. By \cref{claim:multiplicative-weights-mistake-bound}, this implies that the number of mistakes made by $\TransductiveLearner$ is at most $\constUpperBound \sqrt{d}$, as desired.
\end{proof}

    \else
        
    \fi

\hfil

\newpage

\phantomsection

\addcontentsline{toc}{section}{References}
\bibliographystyle{plainnat}
\bibliography{main}

\begin{thebibliography}{25}
\providecommand{\natexlab}[1]{#1}
\providecommand{\url}[1]{\texttt{#1}}
\expandafter\ifx\csname urlstyle\endcsname\relax
  \providecommand{\doi}[1]{doi: #1}\else
  \providecommand{\doi}{doi: \begingroup \urlstyle{rm}\Url}\fi

\bibitem[Balcan and Blum(2010)]{DBLP:journals/jacm/BalcanB10}
Maria{-}Florina Balcan and Avrim Blum.
\newblock A discriminative model for semi-supervised learning.
\newblock \emph{J. {ACM}}, 57\penalty0 (3):\penalty0 19:1--19:46, 2010.
\newblock \doi{10.1145/1706591.1706599}.
\newblock URL \url{https://doi.org/10.1145/1706591.1706599}.

\bibitem[Ben{-}David et~al.(1995)Ben{-}David, Kushilevitz, and Mansour]{DBLP:conf/eurocolt/Ben-DavidKM95}
Shai Ben{-}David, Eyal Kushilevitz, and Yishay Mansour.
\newblock Online learning versus offline learning.
\newblock In Paul M.~B. Vit{\'{a}}nyi, editor, \emph{Computational Learning Theory, Second European Conference, EuroCOLT '95, Barcelona, Spain, March 13-15, 1995, Proceedings}, volume 904 of \emph{Lecture Notes in Computer Science}, pages 38--52. Springer, 1995.
\newblock \doi{10.1007/3-540-59119-2\_167}.
\newblock URL \url{https://doi.org/10.1007/3-540-59119-2\_167}.

\bibitem[Ben{-}David et~al.(1997)Ben{-}David, Kushilevitz, and Mansour]{DBLP:journals/ml/Ben-DavidKM97}
Shai Ben{-}David, Eyal Kushilevitz, and Yishay Mansour.
\newblock Online learning versus offline learning.
\newblock \emph{Mach. Learn.}, 29\penalty0 (1):\penalty0 45--63, 1997.
\newblock \doi{10.1023/A:1007465907571}.
\newblock URL \url{https://doi.org/10.1023/A:1007465907571}.

\bibitem[Ben{-}David et~al.(2008)Ben{-}David, Lu, P{\'{a}}l, and Sot{\'{a}}kov{\'{a}}]{DBLP:journals/corr/abs-0805-2891}
Shai Ben{-}David, Tyler Lu, D{\'{a}}vid P{\'{a}}l, and Miroslava Sot{\'{a}}kov{\'{a}}.
\newblock Learning low-density separators.
\newblock \emph{CoRR}, abs/0805.2891, 2008.
\newblock URL \url{http://arxiv.org/abs/0805.2891}.

\bibitem[Benedek and Itai(1991)]{DBLP:journals/tcs/BenedekI91}
Gyora~M. Benedek and Alon Itai.
\newblock Learnability with respect to fixed distributions.
\newblock \emph{Theor. Comput. Sci.}, 86\penalty0 (2):\penalty0 377--390, 1991.
\newblock \doi{10.1016/0304-3975(91)90026-X}.
\newblock URL \url{https://doi.org/10.1016/0304-3975(91)90026-X}.

\bibitem[Blum and Mitchell(1998)]{DBLP:conf/colt/BlumM98}
Avrim Blum and Tom~M. Mitchell.
\newblock Combining labeled and unlabeled data with co-training.
\newblock In Peter~L. Bartlett and Yishay Mansour, editors, \emph{Proceedings of the Eleventh Annual Conference on Computational Learning Theory, {COLT} 1998, Madison, Wisconsin, USA, July 24-26, 1998}, pages 92--100. {ACM}, 1998.
\newblock \doi{10.1145/279943.279962}.
\newblock URL \url{https://doi.org/10.1145/279943.279962}.

\bibitem[Bousquet et~al.(2021)Bousquet, Hanneke, Moran, van Handel, and Yehudayoff]{DBLP:conf/stoc/BousquetHMHY21}
Olivier Bousquet, Steve Hanneke, Shay Moran, Ramon van Handel, and Amir Yehudayoff.
\newblock A theory of universal learning.
\newblock In Samir Khuller and Virginia~Vassilevska Williams, editors, \emph{{STOC} 2021: 53rd Annual {ACM} {SIGACT} Symposium on Theory of Computing, Virtual Event, Italy, June 21-25, 2021}, pages 532--541. {ACM}, 2021.
\newblock \doi{10.1145/3406325.3451087}.
\newblock URL \url{https://doi.org/10.1145/3406325.3451087}.

\bibitem[Cesa{-}Bianchi and Shamir(2013)]{DBLP:conf/birthday/Cesa-BianchiS13}
Nicol{\`{o}} Cesa{-}Bianchi and Ohad Shamir.
\newblock Efficient transductive online learning via randomized rounding.
\newblock In Bernhard Sch{\"{o}}lkopf, Zhiyuan Luo, and Vladimir Vovk, editors, \emph{Empirical Inference - Festschrift in Honor of Vladimir N. Vapnik}, pages 177--194. Springer, 2013.
\newblock \doi{10.1007/978-3-642-41136-6\_16}.
\newblock URL \url{https://doi.org/10.1007/978-3-642-41136-6\_16}.

\bibitem[Chapelle et~al.(1999)Chapelle, Vapnik, and Weston]{DBLP:conf/nips/ChapelleVW99}
Olivier Chapelle, Vladimir~N. Vapnik, and Jason Weston.
\newblock Transductive inference for estimating values of functions.
\newblock In Sara~A. Solla, Todd~K. Leen, and Klaus{-}Robert M{\"{u}}ller, editors, \emph{Advances in Neural Information Processing Systems 12, {[NIPS} Conference, Denver, Colorado, USA, November 29 - December 4, 1999]}, pages 421--427. The {MIT} Press, 1999.
\newblock URL \url{http://papers.nips.cc/paper/1699-transductive-inference-for-estimating-values-of-functions}.

\bibitem[Chapelle et~al.(2006)Chapelle, Sch{\"{o}}lkopf, and Zien]{DBLP:books/mit/06/CSZ2006}
Olivier Chapelle, Bernhard Sch{\"{o}}lkopf, and Alexander Zien, editors.
\newblock \emph{Semi-Supervised Learning}.
\newblock The {MIT} Press, 2006.
\newblock ISBN 9780262033589.
\newblock \doi{10.7551/MITPRESS/9780262033589.001.0001}.
\newblock URL \url{https://doi.org/10.7551/mitpress/9780262033589.001.0001}.

\bibitem[Darnst{\"{a}}dt et~al.(2013)Darnst{\"{a}}dt, Simon, and Sz{\"{o}}r{\'{e}}nyi]{DBLP:conf/stacs/DarnstadtSS13}
Malte Darnst{\"{a}}dt, Hans~Ulrich Simon, and Bal{\'{a}}zs Sz{\"{o}}r{\'{e}}nyi.
\newblock Unlabeled data does provably help.
\newblock In Natacha Portier and Thomas Wilke, editors, \emph{30th International Symposium on Theoretical Aspects of Computer Science, {STACS} 2013, February 27 - March 2, 2013, Kiel, Germany}, volume~20 of \emph{LIPIcs}, pages 185--196. Schloss Dagstuhl - Leibniz-Zentrum f{\"{u}}r Informatik, 2013.
\newblock \doi{10.4230/LIPICS.STACS.2013.185}.
\newblock URL \url{https://doi.org/10.4230/LIPIcs.STACS.2013.185}.

\bibitem[Gammerman et~al.(1998)Gammerman, Vovk, and Vapnik]{DBLP:conf/uai/GammermanAV98}
Alexander Gammerman, Volodya Vovk, and Vladimir~N. Vapnik.
\newblock Learning by transduction.
\newblock In Gregory~F. Cooper and Seraf{\'{\i}}n Moral, editors, \emph{{UAI} 1998: Proceedings of the Fourteenth Conference on Uncertainty in Artificial Intelligence, University of Wisconsin Business School, Madison, Wisconsin, USA, July 24-26, 1998}, pages 148--155. Morgan Kaufmann, 1998.
\newblock URL \url{https://dslpitt.org/uai/displayArticleDetails.jsp?mmnu=1\&smnu=2\&article\_id=243\&proceeding\_id=14}.

\bibitem[G{\"{o}}pfert et~al.(2019)G{\"{o}}pfert, Ben{-}David, Bousquet, Gelly, Tolstikhin, and Urner]{DBLP:conf/colt/GopfertBBGTU19}
Christina G{\"{o}}pfert, Shai Ben{-}David, Olivier Bousquet, Sylvain Gelly, Ilya~O. Tolstikhin, and Ruth Urner.
\newblock When can unlabeled data improve the learning rate?
\newblock In Alina Beygelzimer and Daniel Hsu, editors, \emph{Conference on Learning Theory, {COLT} 2019, 25-28 June 2019, Phoenix, AZ, {USA}}, volume~99 of \emph{Proceedings of Machine Learning Research}, pages 1500--1518. {PMLR}, 2019.
\newblock URL \url{http://proceedings.mlr.press/v99/gopfert19a.html}.

\bibitem[Hanneke et~al.(2023)Hanneke, Moran, and Shafer]{DBLP:conf/nips/HannekeMS23}
Steve Hanneke, Shay Moran, and Jonathan Shafer.
\newblock A trichotomy for transductive online learning.
\newblock In Alice Oh, Tristan Naumann, Amir Globerson, Kate Saenko, Moritz Hardt, and Sergey Levine, editors, \emph{Advances in Neural Information Processing Systems 36: Annual Conference on Neural Information Processing Systems 2023, NeurIPS 2023, New Orleans, LA, USA, December 10 - 16, 2023}, 2023.
\newblock URL \url{http://papers.nips.cc/paper\_files/paper/2023/hash/3e32af2df2cd13dfbcbe6e8d38111068-Abstract-Conference.html}.

\bibitem[Hanneke et~al.(2024)Hanneke, Raman, Shaeiri, and Subedi]{DBLP:conf/nips/HannekeRSS24}
Steve Hanneke, Vinod Raman, Amirreza Shaeiri, and Unique Subedi.
\newblock Multiclass transductive online learning.
\newblock In Amir Globersons, Lester Mackey, Danielle Belgrave, Angela Fan, Ulrich Paquet, Jakub~M. Tomczak, and Cheng Zhang, editors, \emph{Advances in Neural Information Processing Systems 38: Annual Conference on Neural Information Processing Systems 2024, NeurIPS 2024, Vancouver, BC, Canada, December 10 - 15, 2024}, 2024.
\newblock URL \url{http://papers.nips.cc/paper\_files/paper/2024/hash/6f244818d72b2a4be9b1225d1344e950-Abstract-Conference.html}.

\bibitem[Hoi et~al.(2021)Hoi, Sahoo, Lu, and Zhao]{DBLP:journals/ijon/HoiSLZ21}
Steven C.~H. Hoi, Doyen Sahoo, Jing Lu, and Peilin Zhao.
\newblock Online learning: {A} comprehensive survey.
\newblock \emph{Neurocomputing}, 459:\penalty0 249--289, 2021.
\newblock \doi{10.1016/J.NEUCOM.2021.04.112}.
\newblock URL \url{https://doi.org/10.1016/j.neucom.2021.04.112}.

\bibitem[Joachims(1999)]{DBLP:conf/icml/Joachims99}
Thorsten Joachims.
\newblock Transductive inference for text classification using support vector machines.
\newblock In Ivan Bratko and Saso Dzeroski, editors, \emph{Proceedings of the Sixteenth International Conference on Machine Learning {(ICML} 1999), Bled, Slovenia, June 27 - 30, 1999}, pages 200--209. Morgan Kaufmann, 1999.

\bibitem[Kakade and Kalai(2005)]{DBLP:conf/nips/KakadeK05}
Sham~M. Kakade and Adam Kalai.
\newblock From batch to transductive online learning.
\newblock In \emph{Advances in Neural Information Processing Systems 18 [Neural Information Processing Systems, {NIPS} 2005, December 5-8, 2005, Vancouver, British Columbia, Canada]}, pages 611--618, 2005.
\newblock URL \url{https://proceedings.neurips.cc/paper/2005/hash/17693c91d9204b7a7646284bb3adb603-Abstract.html}.

\bibitem[Littlestone(1987)]{DBLP:journals/ml/Littlestone87}
Nick Littlestone.
\newblock Learning quickly when irrelevant attributes abound: {A} new linear-threshold algorithm.
\newblock \emph{Mach. Learn.}, 2\penalty0 (4):\penalty0 285--318, 1987.
\newblock \doi{10.1007/BF00116827}.
\newblock URL \url{https://doi.org/10.1007/BF00116827}.

\bibitem[Shalev{-}Shwartz and Ben{-}David(2014)]{DBLP:books/daglib/0033642}
Shai Shalev{-}Shwartz and Shai Ben{-}David.
\newblock \emph{Understanding Machine Learning: From Theory to Algorithms}.
\newblock Cambridge University Press, 2014.
\newblock ISBN 978-1-10-705713-5.
\newblock URL \url{http://www.cambridge.org/de/academic/subjects/computer-science/pattern-recognition-and-machine-learning/understanding-machine-learning-theory-algorithms}.

\bibitem[Vapnik(1979)]{vapnik1979estimation}
Vladimir~N. Vapnik.
\newblock \emph{Estimation of Dependencies Based on Empirical Data}.
\newblock Nauka, Moscow, 1979.
\newblock URL \url{https://www.ipu.ru/node/63854/publications}.
\newblock In Russian.

\bibitem[Vapnik(2006)]{DBLP:books/sp/Vapnik06}
Vladimir~N. Vapnik.
\newblock \emph{Estimation of Dependences Based on Empirical Data}.
\newblock Springer, 2nd edition, 2006.
\newblock ISBN 978-0-387-30865-4.
\newblock \doi{10.1007/0-387-34239-7}.
\newblock URL \url{https://doi.org/10.1007/0-387-34239-7}.

\bibitem[Zhu(2005)]{zhu2005semi}
Xiaojin Zhu.
\newblock Semi-supervised learning literature survey.
\newblock Technical report, Department of Computer Sciences, University of Wisconsin--Madison, 2005.

\bibitem[Zhu(2010)]{DBLP:reference/ml/Zhu10}
Xiaojin Zhu.
\newblock Semi-supervised learning.
\newblock In Claude Sammut and Geoffrey~I. Webb, editors, \emph{Encyclopedia of Machine Learning}, pages 892--897. Springer, 2010.
\newblock \doi{10.1007/978-0-387-30164-8\_749}.
\newblock URL \url{https://doi.org/10.1007/978-0-387-30164-8\_749}.

\bibitem[Zhu and Goldberg(2009)]{DBLP:series/synthesis/2009Zhu}
Xiaojin Zhu and Andrew~B. Goldberg.
\newblock \emph{Introduction to Semi-Supervised Learning}.
\newblock Synthesis Lectures on Artificial Intelligence and Machine Learning. Morgan {\&} Claypool Publishers, 2009.
\newblock ISBN 978-3-031-00420-9.
\newblock \doi{10.2200/S00196ED1V01Y200906AIM006}.
\newblock URL \url{https://doi.org/10.2200/S00196ED1V01Y200906AIM006}.

\end{thebibliography}

\appendix

\phantomsection

\addcontentsline{toc}{section}{Appendices}

    \ifdraftcompile
        \section{Halving}

\begin{fact}
    \label{claim:halving}
    Let $\cX$ be a set, and let $\cH \subseteq \zo^{\cX}$ be a hypothesis class. Then for all $n \in \bbN$, all sequences $x \in \cX^n$, and all realizable adversaries, $\Halving$ (\cref{algorithm:halving}) makes at most $\log |\cH|$ mistakes in the transductive online learning (\cref{game:transductive}).\footnote{
        With the suitable syntactic modification, it also makes at most $\log |\cH|$ mistakes in the standard online learning (\cref{game:standard}).
    } Namely,
    \[
    \sup_{n \in \bbN}
    \:
    \sup_{\adversary \in \adversaries_n}
    \:
    \transductivemistakes{\cH,n,\Halving,\adversary}
    \leq \log |\cH|.
    \]
\end{fact}

    \else
        
    \fi

\end{document}